\documentclass[sigconf]{acmart}

\usepackage{mathtools}
\usepackage{bbold}
\usepackage{multirow}
\usepackage[capitalize]{cleveref}
\usepackage{stfloats}

\copyrightyear{2026}
\acmYear{2026}
\setcopyright{cc}
\setcctype{by}
\acmConference[MM '26]{Proceedings of the 34th ACM International Conference on
Multimedia}{November 10--14, 2026}{Rio de Janeiro, Brazil}
\acmBooktitle{Proceedings of the 34th ACM International Conference on Multimedia (MM '26),
November 10--14, 2026, Rio de Janeiro, Brazil}
\acmDOI{10.1145/3767308.3836080}
\acmISBN{979-8-4007-2213-4/2026/11}
\begin{document}

\title{VCAR: Training-Free 3DGS Segmentation via View Completeness and Axis-Aware Boundary
Refinement}

\author{Kun Cao}
\orcid{0009-0004-5078-8681}
\affiliation{%
  \institution{Jinan University}
  \city{Zhuhai}
  \country{China}
}
\email{caokun0526@stu2024.jnu.edu.cn}

\author{Di Wang}
\correspondingauthor
\orcid{0000-0002-2705-731X}
\affiliation{%
  \institution{Jinan University}
  \city{Zhuhai}
  \country{China}
}
\email{diwang@jnu.edu.cn}

\author{Haibin Zhu}
\orcid{0009-0004-2696-2926}
\affiliation{%
  \institution{Jinan University}
  \city{Zhuhai}
  \country{China}
}
\email{zhb2025@stu2025.jnu.edu.cn}

\author{Haozhi Huang}
\orcid{0000-0002-5721-8057}
\affiliation{%
  \institution{Jinan University}
  \city{Zhuhai}
  \country{China}
}
\email{hzhuang@jnu.edu.cn}

\author{Xu Wang}
\orcid{0000-0002-7127-0839}
\affiliation{%
  \institution{Jinan University}
  \city{Zhuhai}
  \country{China}
}
\email{wangxucoco@live.com}

\author{Zheng Shi}
\orcid{0000-0003-3531-9217}
\affiliation{%
  \institution{Jinan University}
  \city{Zhuhai}
  \country{China}
}
\email{zhengshi@jnu.edu.cn}

\author{Guanghua Yang}
\correspondingauthor
\orcid{0000-0002-1598-3033}
\affiliation{%
  \institution{Jinan University}
  \city{Zhuhai}
  \country{China}
}
\email{ghyang@jnu.edu.cn}

\renewcommand{\shortauthors}{Kun Cao et al.}

\begin{abstract}
  Semantic segmentation in 3D Gaussian Splatting (3DGS) is crucial for advancing 3D scene
  understanding.
  Existing methods predominantly rely on feature distillation, which incurs substantial
  per-scene training overhead and often yields blurred segmentation boundaries.
  We identify that these boundary artifacts are driven in part by insufficient
  viewpoint coverage and boundary overflow of anisotropic Gaussian primitives.
  To address these challenges, we propose VCAR, a training-free coarse-to-fine segmentation
  strategy based on View Completeness and Axis-aware Boundary Refinement.
  In the coarse stage, a visibility-based weighted multi-view voting scheme rapidly localizes
  the target.
  In the fine stage, an object-centric sphere derived from the coarse result generates
  supplementary viewpoints via Spherical Spiral Sampling (SSS), allowing multi-view voting on
  the augmented views to precisely refine object boundaries and suppress irrelevant 3D
  Gaussians.
  Moreover, we introduce Axis-aware Boundary Refinement (ABR) to mitigate artifacts from
  anisotropic primitives.
  By decomposing the projected 2D covariance into per-axis contributions, ABR identifies the
  dominant axis responsible for boundary leakage and applies targeted anisotropic compression
  exclusively along that axis.
  Extensive experiments on NVOS and LERF demonstrate that VCAR achieves state-of-the-art
  segmentation accuracy and efficiency without training.
  Our code is available at \url{https://github.com/DDKK0526/VCAR}.
\end{abstract}



\begin{teaserfigure}
  \centering
  \includegraphics[width=0.81\textwidth]{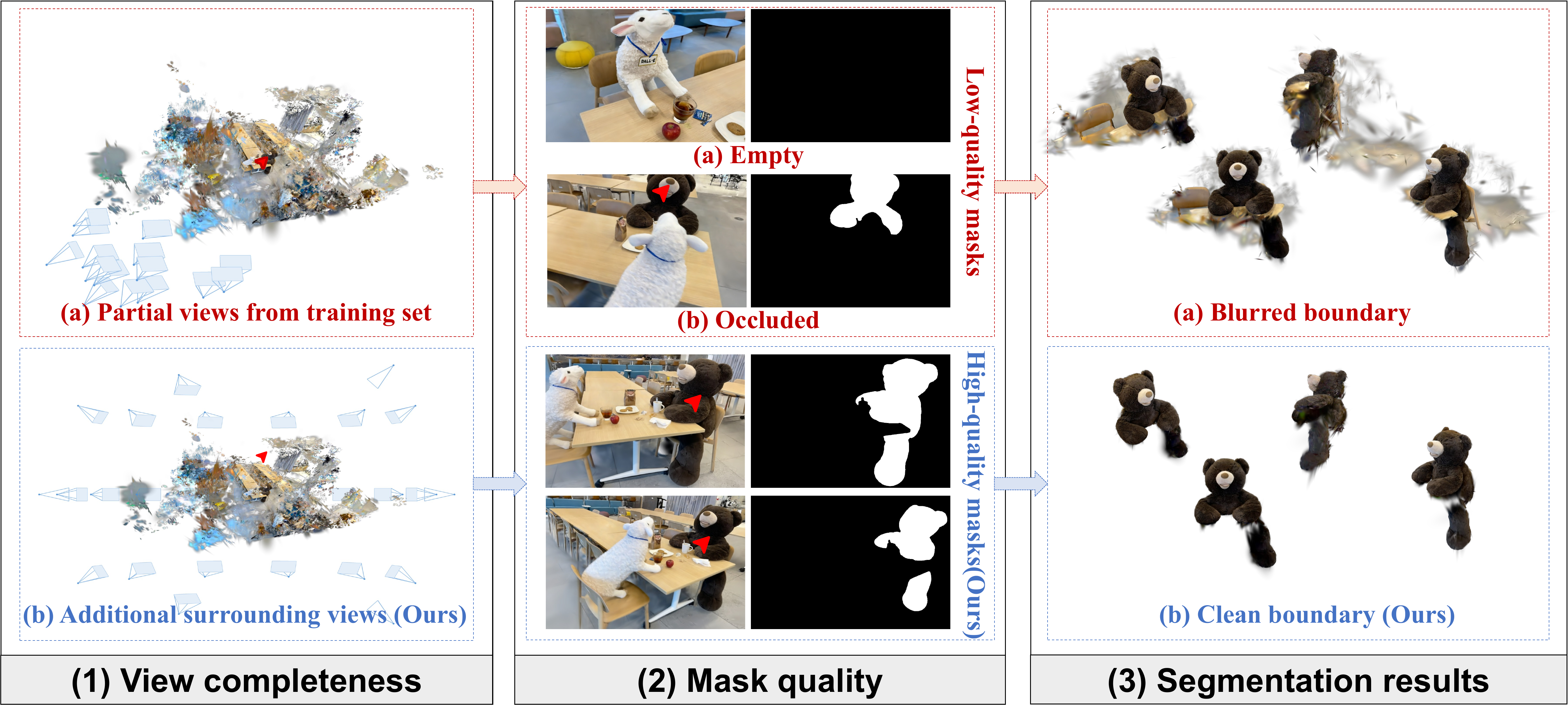}
  \caption{Common problems and comparative results in 3DGS segmentation.}
  \Description{Three comparisons illustrate incomplete training-view coverage, object occlusion
  that degrades mask quality, and blurred 3D segmentation boundaries. Additional views and the
  proposed VCAR method improve coverage, masks, and object boundaries.}
  \label{fig:1}
\end{teaserfigure}

\maketitle

\section{Introduction}
\label{sec:intro}
Understanding and interacting with 3D scenes remains a key challenge in computer vision and
graphics, extending beyond reconstruction to accurate perception and segmentation. Among recent
advances in explicit scene representation, 3D Gaussian Splatting (3DGS) \cite{3DGS} has emerged
as a groundbreaking technique, offering high-fidelity reconstruction and significantly faster
real-time rendering compared to Neural Radiance Fields (NeRF) \cite{Nerf}.

Existing 3DGS segmentation methods \cite{3DGSApplicationSurvey} primarily follow a feature
distillation paradigm: they first use 2D foundation models (e.g., SAM
\cite{kirillov2023segment}, CLIP \cite{CLIP}, DINO \cite{DINOV2}) to generate semantic features
or segmentation masks across multiple views, then distill these 2D features into 3D Gaussian
representations through additional training to embed semantic information into each Gaussian
primitive. While effective for open-vocabulary scene understanding, these approaches rely on
per-scene feature optimization, introducing substantial computational overhead \cite{langsplat,
feature3dgs}---often requiring tens of minutes to hours of additional optimization per scene
before inference. Furthermore, because feature distillation optimizes semantic embeddings via
loss functions across all Gaussians, non-target primitives near object boundaries inevitably
absorb similar semantic features, causing semantic ambiguity between foreground and nearby
background Gaussians \cite{SAGA,marrie2025ludvig}. During segmentation, these semantically
contaminated primitives are erroneously included as part of the target, resulting in blurred
boundaries and floating Gaussian fragments around the object surface.

As illustrated in Figure~\ref{fig:1}, we attribute these boundary artifacts to two
important geometric contributors: insufficient viewpoint coverage and boundary
overflow from anisotropic Gaussians.
Existing methods typically perform feature distillation only from the limited viewpoints in the
training set, whose distribution is often biased and fails to provide uniform coverage of the
target object's surface. When certain regions lack sufficient viewpoint coverage, the Gaussian
primitives in those areas receive inadequate semantic constraints, leading to boundary
ambiguity.
Moreover, the anisotropic ellipsoidal shape of 3D Gaussians causes boundary primitives to extend
beyond the true object surface in their 2D projections, resulting in aliased edges and floating
fragments. Existing methods either ignore this artifact or apply isotropic compression that
indiscriminately shrinks all scale axes, failing to isolate the specific axis responsible for
overflow in a given view.

Based on the above analysis, this paper presents VCAR, a training-free coarse-to-fine
segmentation framework for 3DGS based on View Completeness and Axis-aware Boundary Refinement.
In the coarse stage, VCAR obtains 2D masks from training views via SAM\,3~\cite{SAM3} and
aggregates them through visibility-based weighted voting to rapidly localize the target. In the
fine stage, the coarse result serves as a spatial prior to construct an object-centric sampling
sphere, from which supplementary viewpoints are generated via Spherical Spiral Sampling (SSS)
with uniform and temporally coher
ent coverage.
By rendering only the coarsely segmented Gaussians, inter-object occlusion is
reduced, and visibility-based weighted voting on the augmented viewpoint set
yields a substantially refined segmentation. To further address geometric boundary artifacts, we
propose Axis-aware Boundary Refinement (ABR), which identifies the dominant 3D axis responsible
for boundary overflow and applies targeted anisotropic compression.
The entire framework operates in a purely inference-based manner without any training.
The main contributions are summarized as follows.
\begin{itemize}
  \setlength{\emergencystretch}{1em}
  \item A training-free coarse-to-fine segmentation framework, VCAR, is proposed to
  mitigate two important geometric causes to blurred segmentation boundaries in
  3DGS: insufficient view coverage and anisotropic boundary overflow.
  \item A Spherical Spiral Sampling (SSS) strategy is designed to generate supplementary
  viewpoints for view completeness enhancement, enabling more accurate visibility-based weighted
  voting.
  \item An Axis-aware Boundary Refinement (ABR) method is proposed, in which the scale axis
  responsible for observed 2D boundary overflow is identified, and the Gaussian primitive is
  compressed only along that axis, preserving geometry in other directions.
  \item Extensive experiments on the NVOS and LERF datasets demonstrate that VCAR achieves
  state-of-the-art accuracy and efficiency.
\end{itemize}

\section{Related Work}

\begin{figure*}[!t]
  \centering
  \includegraphics[width=1.0\textwidth]{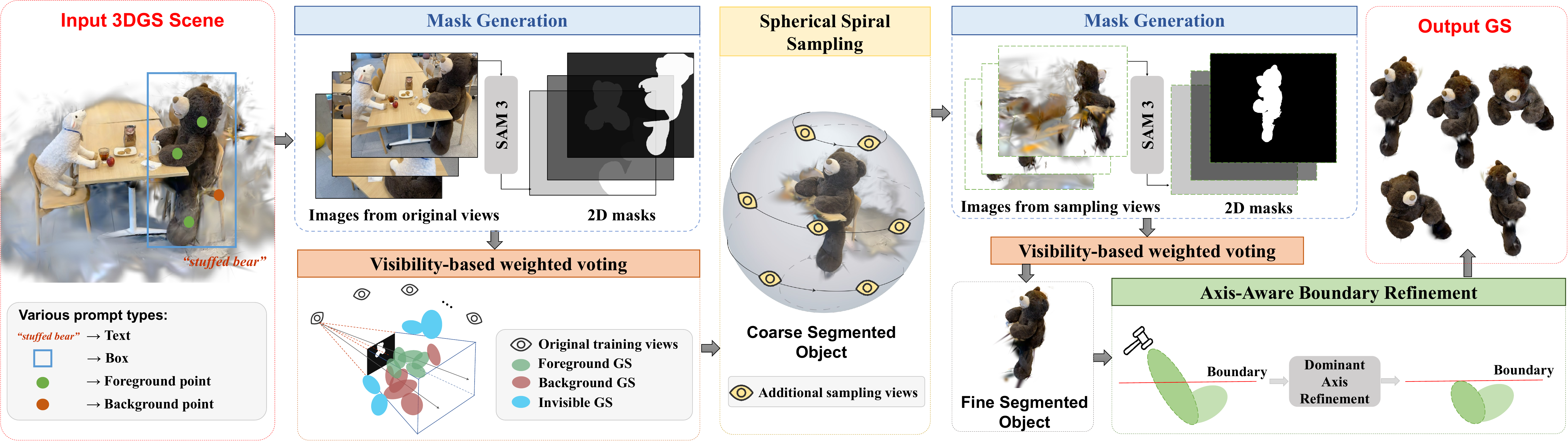}
  \caption{Overview of the VCAR framework. In the coarse stage, original training views are
  segmented and aggregated via visibility-based voting. In the fine stage, only the coarsely
  segmented Gaussians are rendered to eliminate inter-object occlusion, and an object-centric
  sphere generates additional sampling viewpoints via SSS, followed by refined voting and ABR.}
  \Description{A pipeline diagram divided into two stages. The coarse stage shows training-view
  images passing through SAM3 segmentation and visibility-based voting to produce an initial
  Gaussian subset. The fine stage shows an object-centric sphere with a spiral trajectory
  generating supplementary camera viewpoints, followed by refined voting and an Axis-aware
  Boundary Refinement module that refines the overflowing 3D axis of boundary Gaussians.}
  \label{fig:framework}
\end{figure*}

\subsection{3D Gaussian Splatting}
3D Gaussian Splatting (3DGS)~\cite{3DGS} represents scenes using fully explicit, anisotropic
Gaussian primitives.
Pixel colors are rendered via a highly parallelized tile-based rasterizer that projects the 3D
covariance into 2D and performs front-to-back $\alpha$-blending,
achieving real-time novel view synthesis comparable to NeRF~\cite{Nerf} but with vastly
accelerated optimization.
Its structured nature has inspired extensive follow-up research.
Works have advanced reconstruction quality and
structure~\cite{mipsplatting,Scaffold,gaussianpro,yuan2025robust},
storage efficiency through compaction~\cite{lightgaussian,compact3dgs},
dynamic scene modeling~\cite{gaufre,4dgs}, and generalization from sparse
views~\cite{pixelsplat,fsgs}.
Beyond reconstruction, 3DGS has been extended to downstream applications like 3D
editing~\cite{Gaussianeditor,qiu2024language,physgaussian} and
generation~\cite{luciddreamer,dreamgaussian}.
More recently, semantic understanding and interactive segmentation have emerged as an active
frontier.
Notably, while the anisotropic ellipsoidal geometry of Gaussians excels at capturing surface
details,
it also introduces segmentation challenges: boundary Gaussians routinely protrude beyond true
object surfaces in 2D projections, producing aliased edges.
This fundamental geometric artifact motivates the boundary refinement proposed in this work.

\subsection{3D Segmentation in 3DGS}
Recent advances in 2D foundation models (\emph{e.g.}, SAM~\cite{kirillov2023segment},
CLIP~\cite{CLIP}, DINO~\cite{DINOV2}) have greatly enhanced 3D scene perception.
Transferring such 2D semantic capabilities into 3DGS representations has become a core pathway
for 3D segmentation,
and existing methods can be grouped into three categories based on how semantic information is
incorporated.

\textbf{Feature Distillation-based Methods.}
These methods~\cite{kerr2023lerf,langsplat,langsplatv2,%
  feature3dgs,shi2024language,gao2024fast,chen2025slgaussian}
distill 2D semantic knowledge into 3D Gaussian representations.
LangSplat~\cite{langsplat} trains per-Gaussian language features via a scene-specific
autoencoder supervised by CLIP embeddings extracted from SAM-generated hierarchical masks, while
Feature3DGS~\cite{feature3dgs} distills LSeg~\cite{lseg} and SAM features and leverages SAM's
decoder for 2D interpretation.
Subsequent works improve along two directions.
On the accuracy side, LangSurf~\cite{li2024langsurf} jointly optimizes language Gaussians on
object surfaces with dense pixel-level geometry supervision,
while OpenGaussian~\cite{wu2024opengaussian} adopts coarse-to-fine feature discretization with
instance-level 3D-2D association.
On the efficiency side, LEGaussian~\cite{shi2024language} quantizes dense language features into
a discrete space to reduce memory overhead,
FMGS~\cite{zuo2025fmgs} integrates multi-resolution hash encodings for efficient semantic
embedding,
and LangSplatV2~\cite{langsplatv2} represents each Gaussian as a sparse code in a global
dictionary, enabling decoder-free sparse coefficient splatting with CUDA-optimized rendering.
Despite significant progress, feature distillation inherently relies on per-scene optimization,
introducing substantial computational overhead.
Moreover, since all Gaussians are jointly optimized, non-target primitives near object
boundaries inevitably absorb similar semantic features,
leading to semantic ambiguity and blurred segmentation boundaries.

\textbf{2D Mask Lifting-based Methods.}
Another line of
work~\cite{gaussiangrouping,lyu2024gaga,ying2024omniseg3d,%
  SAGA,clickgaussian,cobgs,liao2025clipgs}
directly lifts 2D masks into 3D, with multi-view consistency as the central challenge.
GaussianGrouping~\cite{gaussiangrouping} aligns cross-view masks via an object association
technique,
while Gaga~\cite{lyu2024gaga} employs a 3D-aware memory bank to associate masks across diverse
camera poses.
OmniSeg3D~\cite{ying2024omniseg3d} refines lifted masks through hierarchical contrastive
learning and clustering.
SAGA~\cite{SAGA} distills SAM's segmentation capability into a scale-gated affinity
representation,
Click-Gaussian~\cite{clickgaussian} achieves interactive segmentation through multi-granularity
feature fields,
and COB-GS~\cite{cobgs} jointly optimizes masks and textures with boundary-adaptive Gaussian
splitting to refine boundary structures.
While these approaches have improved multi-view consistency, they still require additional
per-scene training
and remain susceptible to boundary artifacts caused by insufficient viewpoint coverage and
anisotropic Gaussian overflow.

\textbf{Training-Free Mask Lifting-based Methods.}
To eliminate training overhead, several recent methods explore training-free
strategies~\cite{hu2024sagd,shen2024flashsplat,jain2024gaussiancut,%
  zhao2025isegman,marrie2025ludvig,zhang2025labelgs,%
  chacko2025lifting,bao2025segwild}.
SAGD~\cite{hu2024sagd} classifies Gaussians by projecting their centers onto 2D masks.
FlashSplat~\cite{shen2024flashsplat} formulates mask lifting as a linear programming problem,
while GaussianCut~\cite{jain2024gaussiancut} applies graph-cut optimization for
foreground-background partitioning.
iSegMan~\cite{zhao2025isegman} introduces a visibility-guided voting scheme weighted by Gaussian
opacity,
and LUDVIG~\cite{marrie2025ludvig} performs inverse feature aggregation with graph diffusion for
feature refinement.
However, these methods primarily focus on cross-view consistency or efficient aggregation,
while paying less attention to two important geometric contributors to boundary
artifacts:
insufficient viewpoint coverage that leaves boundary primitives under-constrained,
and boundary overflow of anisotropic Gaussians that existing methods either ignore entirely or
address with indiscriminate isotropic compression.

\subsection{Boundary Refinement in 3DGS Segmentation}
Resolving the boundary ambiguity caused by the volumetric nature of Gaussians
remains a critical challenge.
As noted in the preceding paragraphs, SAGD~\cite{hu2024sagd} and COB-GS~\cite{cobgs}
also address this issue from a structural perspective:
they identify boundary Gaussians that span both foreground and background regions
and apply physical splitting or decomposition to separate them.
GaussianTrimmer~\cite{liao2026gaussiantrimmer} takes an alternative route,
introducing an online post-processing step that renders virtual views
to actively trim primitives overflowing 2D boundaries.
LBG~\cite{chacko2025lifting} circumvents structural modification entirely
by anchoring each 2D pixel strictly to the primitive providing the maximum alpha-blending
weight,
thereby avoiding semantic mixing.
However, splitting and trimming operations inherently impose destructive modifications
that compromise the anisotropic geometric properties of the optimized 3DGS scene.
Furthermore, prior solutions generally employ isotropic compression
that indiscriminately compresses all spatial dimensions.

In contrast, VCAR jointly targets both insufficient viewpoint coverage
and boundary overflow within a unified, training-free framework.
By introducing object-centric multi-view augmentation,
we provide boundary primitives with additional angular constraints.
More importantly, our proposed ABR is the first to trace boundary overflow
back to specific 3D scale axes,
applying axis-selective anisotropic compression along the culpable axis
rather than indiscriminate isotropic compression.
This better preserves the structural fidelity of well-behaved directions
while improving boundaries conformity,
simultaneously eliminating per-scene training overhead
and achieving high-precision segmentation.

\section{Method}

\subsection{Overview}
\label{sec:overview}

Given a 3DGS scene with $N$ Gaussian primitives $\mathcal{G} = \{g_i\}_{i=1}^N$,
where each $g_i$ is parameterized by position $\boldsymbol{\mu}_i \in \mathbb{R}^3$,
3D covariance $\boldsymbol{\Sigma}_i$, opacity $\alpha_i$,
and spherical harmonics coefficients $\mathbf{\textit{c}}_i$,
along with training viewpoints $\mathcal{V}^{\text{train}} = \{v^{(1)}, \ldots, v^{(M)}\}$,
VCAR identifies the subset $\mathcal{G}^* \subseteq \mathcal{G}$ belonging to a user-specified
target
from a segmentation prompt on a reference view.
The framework adopts a coarse-to-fine strategy without any training (shown in Figure
\ref{fig:framework}):

\textbf{Stage~1: Coarse Segmentation Stage.}
The input 3DGS scene is rendered from training viewpoints and segmented by SAM\,3~\cite{SAM3}
based on user-provided prompts.
We specifically integrate SAM\,3 because it natively supports both geometric and text prompts,
thereby enriching the user's input modalities.
The resulting 2D masks are then aggregated via visibility-based weighted voting
(\cref{sec:voting})
to produce a coarse segmentation result $\mathcal{G}^{\text{coarse}}$.

\textbf{Stage~2: Fine Segmentation Stage.}
Building upon $\mathcal{G}^{\text{coarse}}$,
an object-centric sphere is robustly estimated and supplementary viewpoints are generated via
SSS (\cref{sec:sampling})
to improve angular coverage while maintaining temporal smoothness.
Visibility-based weighted voting (\cref{sec:voting}) is reapplied on the augmented viewpoint
set, which substantially refines the segmentation, and ABR (\cref{sec:boundary}) further
mitigates boundary artifacts to yield the final result $\mathcal{G}^*$.

\subsection{Visibility-based Weighted Voting}
\label{sec:voting}
The $M$ rendered images are sequentially segmented by SAM\,3~\cite{SAM3} operating in video
segmentation mode, yielding per-frame binary masks $\{\mathcal{M}^{(j)}\}_{j=1}^{M}$.
The visibility-based weighted voting mechanism to aggregate the masks
$\{\mathcal{M}^{(j)}\}_{j=1}^{M}$, and then produce the segmented subset
$\mathcal{G}_{\text{seg}}$.

\begin{figure}[t]
    \centering
    \includegraphics[width=0.98\columnwidth]{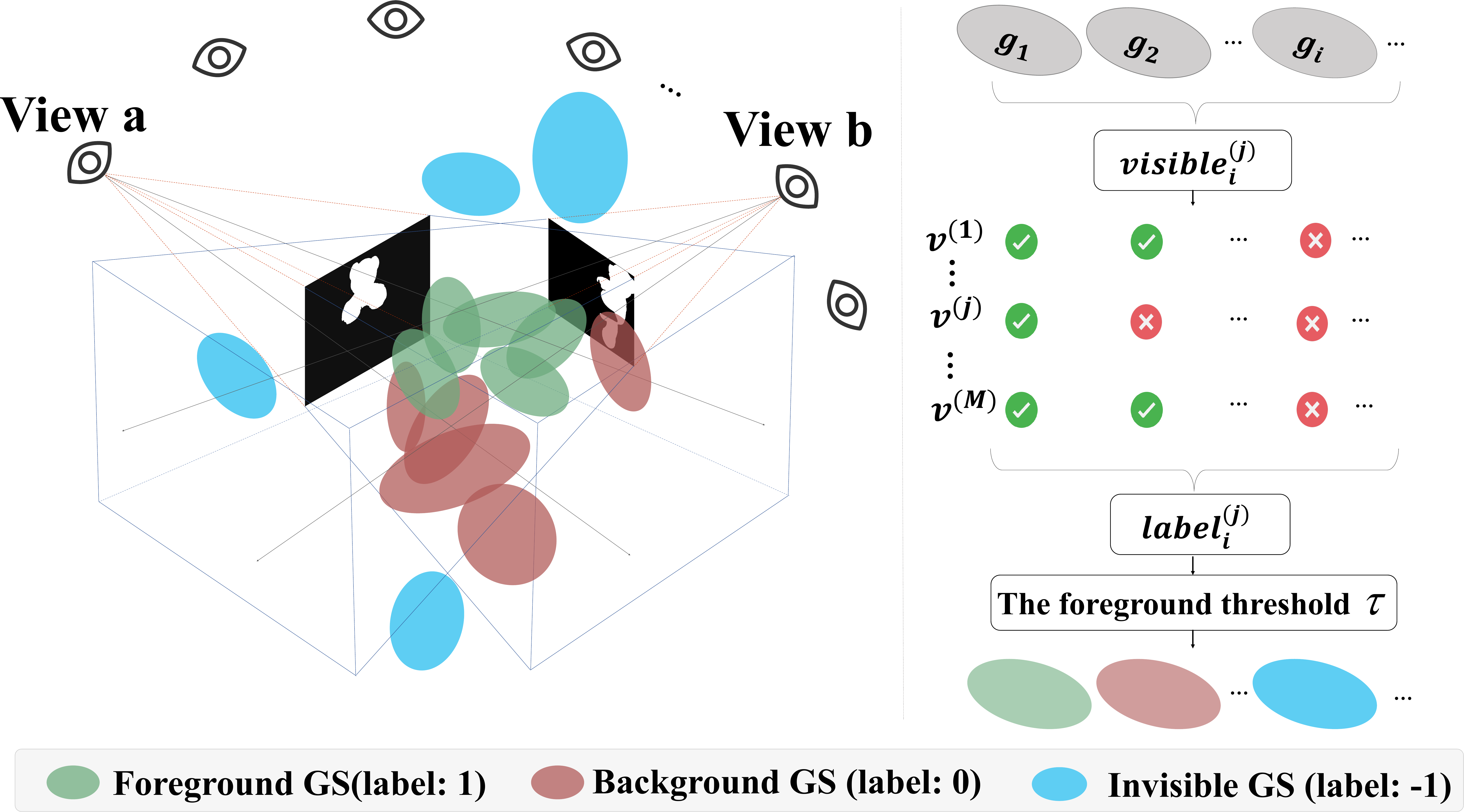}
    \caption{Visibility-based Weighted Voting.}
    \Description{A flow diagram showing per-view Gaussian projection, visibility filtering,
    foreground and background voting, and aggregation into final Gaussian labels.}
    \label{fig:VWV_pipeline}
\end{figure}

Unlike Naive voting, the visibility-based weighted voting mechanism computes each Gaussian
primitive's foreground ratio exclusively over views in which it is visible, shown in Figure
\ref{fig:VWV_pipeline}. Concretely, for the $j$-th viewpoint $v^{(j)}$, the $i$-th Gaussian
center $\boldsymbol{\mu}_i$ is transformed into camera coordinates via the world-to-camera
matrix $\mathbf{W}^{(j)} \in \mathbb{R}^{4 \times 4}$
and projected to pixel coordinates $(u_i^{(j)}, v_i^{(j)})$
through the full projection matrix $\mathbf{P}_j$.
A primitive is deemed visible in view $v^{(j)}$ if its depth $z_i^{(j)} $ is positive and its
projection falls within the image bounds $W_I^{(j)} \times H_I^{(j)}$:
\begin{equation}
  \text{visible}_i^{(j)}= \bigl(z_i^{(j)} > 0\bigr)
  \;\wedge\; \bigl(0 \leq u_i^{(j)} < W_I^{(j)}\bigr) \;\wedge\; \bigl(0 \leq v_i^{(j)} <
  H_I^{(j)}\bigr)
\end{equation}
If visible, $\mathcal{M}^{(j)} \in \{0, 1\}$; if invisible, $\mathcal{M}^{(j)} = -1 $.
Hence, each Gaussian primitive receives a tri-valued label:
\begin{equation}
  label_i^{(j)} = \begin{cases} 1 & \text{foreground} \\ 0 & \text{background}\\ -1 &
  \text{invisible} \end{cases}
\end{equation}
The foreground ratio is then computed only over visible views
and thresholded to determine segmentation:
\begin{equation}
  R_i = \frac{\sum_{j} \mathbb{1}[label_i^{(j)} = 1]}{\max\bigl(\sum_{j}
  \mathbb{1}[label_i^{(j)} \neq -1],\; 1\bigr)},
  \qquad
  y_i = \mathbb{1}[R_i \geq \tau]
\end{equation}
where $\tau$ is the foreground ratio threshold.
Therefore, the segmentation result is
$\mathcal{G}_{\text{seg}} = \{g_i \in \mathcal{G} \mid y_i = 1\}$.
In the proposed VCAR, visibility-based weighted voting is utilized at both coarse and fine
segmentation stages.
In the coarse stage, the input 2D mask is from training views and the coarse segmentation result
is marked as $\mathcal{G}^{\text{coarse}}$. In the fine stage, the input 2D mask is from the
additional sampling views and the fine segmentation result is marked as
$\mathcal{G}^{\text{fine}}$.

\subsection{Spherical Spiral Sampling (SSS)}
\label{sec:sampling}
A lightweight view completeness assessment determines whether the Spherical Spiral Sampling
(SSS) is required: if the existing training views provide sufficient angular coverage of the
target object, SSS is skipped to maximize inference efficiency; otherwise, supplementary
viewpoints are generated using SSS to ensure full coverage.

\subsubsection{Object-Centric Sphere Estimation}
We construct an object-centric bounding sphere from $\mathcal{G}^{\text{coarse}}$,
parameterized by a center $\boldsymbol{c}$ and radius $r$.
Let $\mathcal{P} = \{\boldsymbol{\mu}_i \mid g_i \in \mathcal{G}^{\text{coarse}}\}$ denote the
Gaussian positions
and $\bar{\boldsymbol{\mu}}$ presents their preliminary mean.
We compute each point's distance $d_i = \|\boldsymbol{\mu}_i - \bar{\boldsymbol{\mu}}\|_2$
and retain a robust inlier subset via $3\sigma$ rejection:
\begin{equation}
  \mathcal{P}^* = \bigl\{\boldsymbol{\mu}_i \in \mathcal{P} \mid d_i \leq \mu_d +
  3\sigma_d\bigr\},
  \qquad
  \boldsymbol{c} = \frac{1}{|\mathcal{P}^*|} \sum_{\boldsymbol{\mu}_i \in \mathcal{P}^*}
  \boldsymbol{\mu}_i
\end{equation}
where $\mu_d$ and $\sigma_d$ are the mean and standard deviation of $\{d_i\}$.
The sphere radius is set to
\begin{equation}
  r = \eta \cdot \|v_{\text{ref}}^{\text{pos}} - \boldsymbol{c}\|_2
\end{equation}
where $v_{\text{ref}}^{\text{pos}}$ is the camera position of the view on which the user
provides the segmentation prompt
and $\eta$ is a scaling factor to better ensure that the target is fully contained for generated
viewpoints.
The $3\sigma$ rejection prevents the estimated center from being biased by scattered outlier
primitives retained from the coarse stage.

\subsubsection{View Completeness Assessment}
With the robust object center $\boldsymbol{c}$ established,
we assess view completeness by measuring the maximum angular gap on the unit sphere.
Specifically, we generate $K=2000$ uniformly distributed test directions
$\{\mathbf{t}_i\}_{i=1}^{K}$ via a Fibonacci lattice,
and for each valid training camera compute the unit heading $\mathbf{h}_j$ from its position
toward $\boldsymbol{c}$.
The maximum angular gap is then:
\begin{equation}
\Delta_{\max} = \max_{i} \; \arccos \!\left( \max_{j} \; \mathbf{t}_i^\top \mathbf{h}_j \right)
\end{equation}
If $\Delta_{\max}$ exceeds a threshold $\Delta_{\text{th}}$ (e.g., $90^\circ$), SSS is triggered
to augment coverage; otherwise, it is bypassed for efficiency. See the supplementary material
for detailed camera filtering criteria.

\subsubsection{Spiral Trajectory Sampling}
Additional viewpoints are sampled along a continuous spherical spiral trajectory on the
estimated object-centric sphere.
Specifically, let $N_s = S_1 \times S_2$ be the total number of sampled points,
where $S_1$ is the number of spiral revolutions and $S_2$ is the number of points per
revolution.
For the $k$-th point ($k = 0, \ldots, N_s-1$),
we define the azimuthal and elevation angles:
\begin{equation}
  \theta_k = \frac{2\pi S_1 \cdot k}{N_s}, \qquad
  \phi_k = \phi_{\text{s}} + \frac{(\phi_{\text{e}} - \phi_{\text{s}}) \cdot k}{N_s - 1}
\end{equation}
where $\phi_{\text{s}}$ and $\phi_{\text{e}}$ are the initial and final elevation angles,
respectively.
Supplementary camera positions  are computed via spherical-to-Cartesian conversion:
\begin{equation}
  \mathbf{p}_k = \boldsymbol{c} + r \cdot
  \bigl(\cos\phi_k \cos\theta_k, \; \cos\phi_k \sin\theta_k, \; \sin\phi_k\bigr)
\end{equation}

All cameras are oriented toward the sphere center $\boldsymbol{c}$, collectively forming the
supplementary viewpoint set $\mathcal{V}^{\text{sampled}}$.
The inherent continuity of spiral sampling yields frames with smoothly varying camera poses, a
property that benefits SAM\,3~\cite{SAM3} operating in video segmentation mode, which relies on
temporal coherence across consecutive frames.
Rendering the augmented viewpoint set $\mathcal{V}^{\text{fine}} = \mathcal{V}^{\text{train}}
\cup \mathcal{V}^{\text{sampled}}$ using only $\mathcal{G}^{\text{coarse}}$
reduces inter-object occlusion and helps produce more reliable 2D masks.
The visibility-based weighted voting strategy (\cref{sec:voting})  is then reapplied over
$\mathcal{V}^{\text{fine}}$ to obtain the refined Gaussian subset $\mathcal{G}^{\text{fine}}$.

\subsection{Axis-aware Boundary Refinement (ABR)}
\label{sec:boundary}
\begin{figure}[t]
    \centering
    \includegraphics[width=0.9\columnwidth]{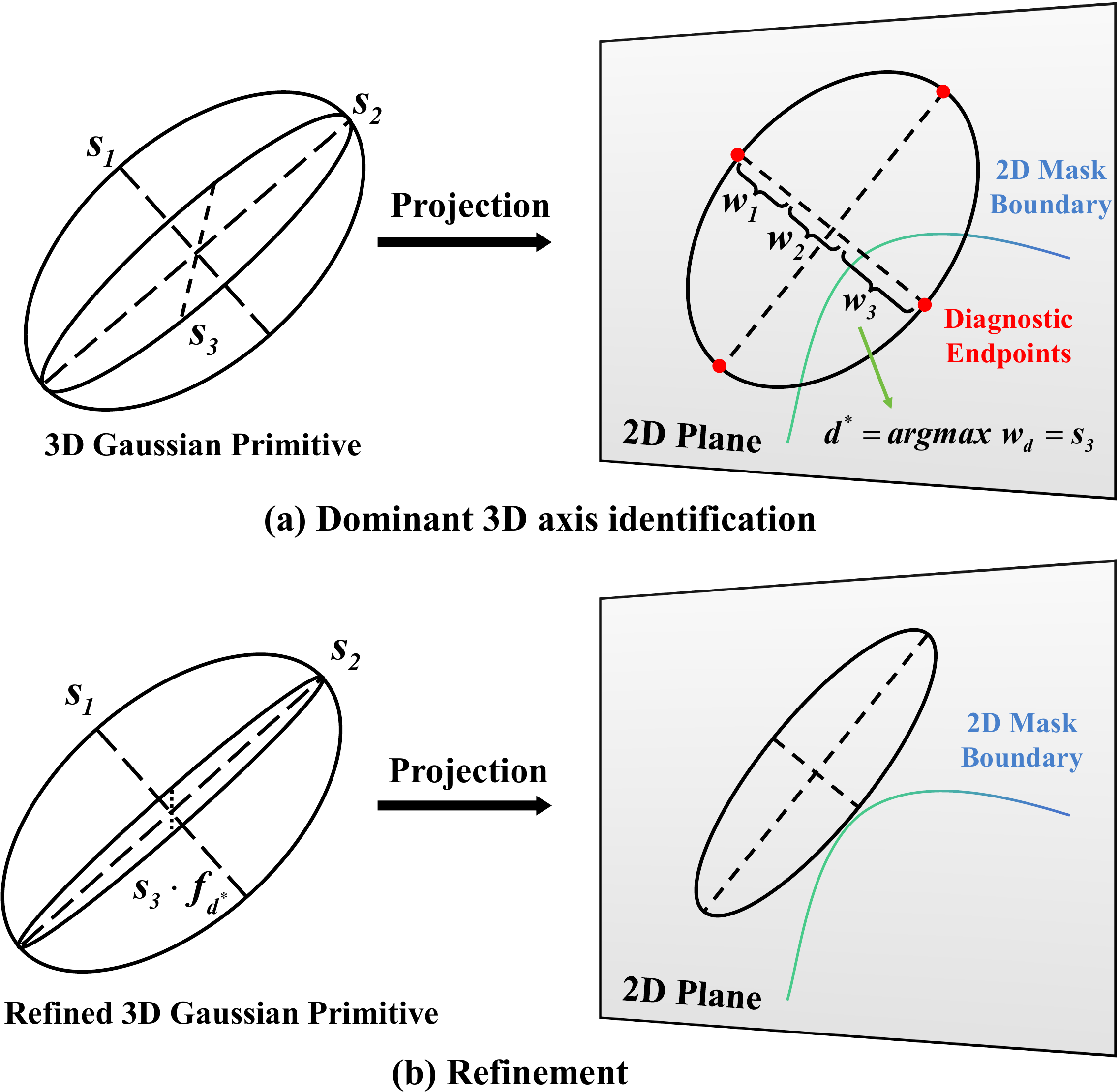}
    \caption{Axis-aware Boundary Refinement.}
    \Description{A diagram showing how projected Gaussian boundary overflow is detected,
    attributed to a dominant three-dimensional scale axis, and corrected by compressing that
    axis.}
    \label{fig:abr_pipeline}
\end{figure}

In $\mathcal{G}^{\text{fine}}$, boundary Gaussians may still protrude beyond the object mask in
2D renderings due to anisotropic scaling. As illustrated in Figure \ref{fig:abr_pipeline}, ABR
detects such boundary overflow from the projected ellipse geometry, traces it to the dominant 3D
axis, and applies targeted per-axis refinement guided by multi-view consistency.
\subsubsection{Boundary Overflow Detection.}
For Gaussian $g_i$ visible in view $v^{(j)}$, splatting yields the projected center
$\boldsymbol{m}_i^{(j)} = (u_i^{(j)},\, v_i^{(j)})$
(see \cref{sec:voting})
and a $2\times2$ covariance matrix $\boldsymbol{\Sigma}_{2D,i}^{(j)}$.
We suppress the view superscript~$(j)$ in the per-view derivations below for brevity.
The principal orientation and scales of the projected ellipse are derived from the
eigendecomposition
$\boldsymbol{\Sigma}_{2D,i} = \mathbf{U}\,\mathrm{diag}(\lambda_1,\lambda_2)\,\mathbf{U}^\top$
with $\lambda_1 \geq \lambda_2$, and the rendering cutoff at $\sigma_c = 3$ standard deviations
gives semi-axis lengths $a_i = \sigma_c\sqrt{\lambda_1}$ and $b_i = \sigma_c\sqrt{\lambda_2}$.
Four diagnostic endpoints are placed at the extremities of both principal axes:
\begin{equation}
  \mathcal{E}_i = \bigl\{\boldsymbol{m}_i \pm a_i\,\vec{e}_1,\;\boldsymbol{m}_i \pm
  b_i\,\vec{e}_2\bigr\}
\end{equation}
where $\vec{e}_1, \vec{e}_2$ are the unit eigenvectors of $\boldsymbol{\Sigma}_{2D,i}$.

A Gaussian primitive is marked as overflowing in view $v^{(j)}$ if any endpoint falls outside
the foreground mask. This directional test preserves elongated Gaussians whose semi-major axis
lies entirely within the mask. To suppress false positives, boundary overflow detections are
aggregated across views: a Gaussian is flagged for compression only if its overflow ratio
$\gamma_i = n_i^{\text{ovf}} / n_i^{\text{vis}}$ exceeds a tolerance threshold $\rho$, where
$n_i^{\text{vis}}$ and $n_i^{\text{ovf}}$ denote the number of views in which the Gaussian
center is visible and in which overflow is detected, respectively.

\subsubsection{Dominant 3D Axis Identification.}
For each flagged Gaussian $g_i$ with rotation matrix $\mathbf{R}_i$ (columns $\mathbf{r}_1,
\mathbf{r}_2, \mathbf{r}_3$ define the local axes) and scale vector $\mathbf{s} = (s_1, s_2,
s_3)$, we trace the observed 2D overflow to its originating 3D scale axis. Under the linearized
splatting model, the 2D covariance admits a per-axis decomposition:
\begin{equation}
  \label{eq:cov_decomp}
  \boldsymbol{\Sigma}_{2D,i}
  \;=\; \mathbf{M}\,\boldsymbol{\Sigma}_i\,\mathbf{M}^\top
  \;=\; \sum_{d=1}^{3} s_d^{\,2}\;\mathbf{q}_d\,\mathbf{q}_d^\top
\end{equation}
where $\mathbf{M} = \mathbf{J}\,\mathbf{W}_{R}^{(j)}$ is the linearized projection matrix.
Specifically,
$\mathbf{W}_{R}^{(j)}$ denotes the $3\times3$ rotation matrix of the world-to-camera matrix
$\mathbf{W}^{(j)}$ (see \cref{sec:voting});
$\mathbf{J}$ represents the perspective Jacobian evaluated at the Gaussian center;
$\mathbf{q}_{d} = \mathbf{M}\,\mathbf{r}_d \in \mathbb{R}^2$
is the projected direction of the $d$-th local axis.
Let $\mathbf{u}$ denote the eigenvector direction
of a specific overflowing endpoint,
i.e., $\mathbf{u} = \vec{e}_1$ for major-axis
and $\mathbf{u} = \vec{e}_2$ for minor-axis overflow,
with corresponding eigenvalue~$\lambda$:
\begin{equation}
  \label{eq:var_decomp}
  \lambda
  \;=\; \mathbf{u}^\top \boldsymbol{\Sigma}_{2D,i}\,\mathbf{u}
  \;=\; \sum_{d=1}^{3}
  \underbrace{s_d^{2}\,(\mathbf{u}^\top\mathbf{q}_d)^2}_{\triangleq\;w_d}
\end{equation}
Where $w_d$ quantifies the variance contribution
of the $d$-th 3D axis along the overflow direction.

A large $w_d$ indicates that the $d$-th axis is both geometrically long and well-aligned with
$\mathbf{u}$ in the projected view.
Conversely, axes nearly parallel to the viewing direction
project to near-zero $\mathbf{q}_d$ and thus contribute negligibly, regardless of their physical
scale.
Based on these contributions, the dominant overflowing axis is identified as $d^* =
\arg\max_{d}\,w_d$.

\subsubsection{Dominant Axis Refinement.}
For each overflowing endpoint,
we trace from the projected center $\boldsymbol{m}_i$
along $\mathbf{u}$ and sample the mask to find
the directional boundary distance $\ell_{\mathbf{u}}$,
defined as the distance from the center to the
mask boundary along $\mathbf{u}$.
The goal is to compress the dominant axis so that the rendered extent along $\mathbf{u}$
matches $\ell_{\mathbf{u}}$ to correct the overflow.
Specifically, we scale axis $d^*$ by a compression factor
$f_{d^*} \in [f_{\min},\,1]$, i.e., $s_{d^*} \to f_{d^*}\,s_{d^*}$.
This replaces $w_{d^*}$ with $f_{d^*}^{\,2}\,w_{d^*}$
while all other terms remain fixed,
giving the post-compression variance along $\mathbf{u}$:
\begin{equation}
  \lambda'
  \;=\; (\lambda - w_{d^*}) \;+\; f_{d^*}^{\,2}\,w_{d^*}
\end{equation}
Setting $\sigma_c\sqrt{\lambda'} = \ell_{\mathbf{u}}$
so that the recalibrated rendering extent matches the boundary distance
and solving for $f_{d^*}$:
\begin{equation}
  \label{eq:compression_factor}
  f_{d^*}
  \;=\; \sqrt{
    \frac{(\ell_{\mathbf{u}}/\sigma_c)^2 \;-\; \lambda \;+\; w_{d^*}}
    {w_{d^*}}
  }
\end{equation}
Since different viewpoints observe different projections of the same Gaussian,
an overflow endpoint in one view may implicate a different 3D axis $d^*$ than in another,
and a single axis may yield varying compression factors across different observations.
To reconcile these multi-view inconsistencies, the final compression factor for each 3D axis~$d$
is chosen as the minimum (i.e., tightest) $f_d$ among all its attributions,
clamped to a lower bound $f_{\min}$.
Taking the minimum conservatively ensures that every observed 2D overflow is fully corrected
across all views.
Finally, since 3DGS parameterizes scales in log-space,
the update is applied additively:
\begin{equation}
  \log s_{d^*} \leftarrow \log s_{d^*} + \log f_{d^*},
\end{equation}
leaving all non-dominant axes untouched to
preserve the Gaussian's geometry along well-behaved directions.

\section{Experiments}
\label{sec:experiments}

\subsection{Experimental Settings}
\label{sec:eval_setup}

\subsubsection{Datasets.}
We evaluate VCAR on two widely adopted 3DGS segmentation benchmarks.
\textbf{NVOS}~\cite{ren2022nvos} provides seven real-world forward-facing scenes
with per-view binary masks, where the limited viewpoint diversity
makes it particularly suitable for evaluating view completeness enhancement.
\textbf{LERF}~\cite{kerr2023lerf} comprises four indoor tabletop scenes
with 85 annotated objects featuring complex inter-object occlusion,
providing a challenging testbed for boundary refinement-level segmentation.

\subsubsection{Metrics.}
Following the prior works \cite{SAGA,hu2024sagd,marrie2025ludvig}, we report \textbf{mean
Intersection over Union (mIoU)} and \textbf{mean pixel Accuracy (mAcc)}.
All metrics are computed by rendering segmented Gaussians from held-out test views and comparing
against ground-truth masks.

\subsection{Implementation Details}
\label{sec:impl_details}

We implement our method using PyTorch~\cite{paszke2019pytorch} and the gsplat rendering
backend~\cite{ye2025gsplat}.
The entire framework is training-free, and all pipeline stages execute at inference time
on a single NVIDIA A100 GPU.
SAM\,3~\cite{SAM3} is adopted as the 2D segmentation backbone, operating in video segmentation
mode to leverage temporal coherence across consecutively rendered frames.

In the coarse stage, training-view images are rendered at original resolution and segmented
using user-provided prompts. For view completeness assessment, we adopt $K = 2000$
Fibonacci-lattice test directions with an angular gap threshold of $\Delta_{\text{th}} =
90^\circ$.
In the fine stage, the sphere radius scaling factor is set to $\eta = 1.2$, and the spherical
spiral is configured with $S_1 = 4$ revolutions and $S_2 = 8$ points per revolution, yielding 32
supplementary viewpoints per object.
The elevation range is defined from $\phi_{\text{s}} = -60^\circ$ to $\phi_{\text{e}} =
60^\circ$.

For NVOS, the voting thresholds are set to $\tau = 0.5$ (coarse stage) and $\tau = 0.8$ (fine
stage). For LERF, objects are generally small and densely arranged, resulting in fewer Gaussian
primitives per object. Therefore, we adopt lower thresholds, $\tau = 0.4$ for coarse and $\tau
\in [0.5,\, 0.7]$ for fine to prioritize target completeness, accepting the inclusion of some
redundant boundary Gaussians.
This design complements ABR, which subsequently removes boundary overflow while preserving
object integrity. For Axis-Aware Boundary Refinement, the rendering cutoff is set to $\sigma_c =
3$ standard deviations, the multi-view overflow tolerance ratio to $\rho = 0.6$,
and the minimum compression factor to $f_{\min} = 0.1$.

\subsection{Quantitative Results}
\label{sec:quant_results}

\begin{figure*}[t]
    \centering
    \includegraphics[width=0.88\textwidth]{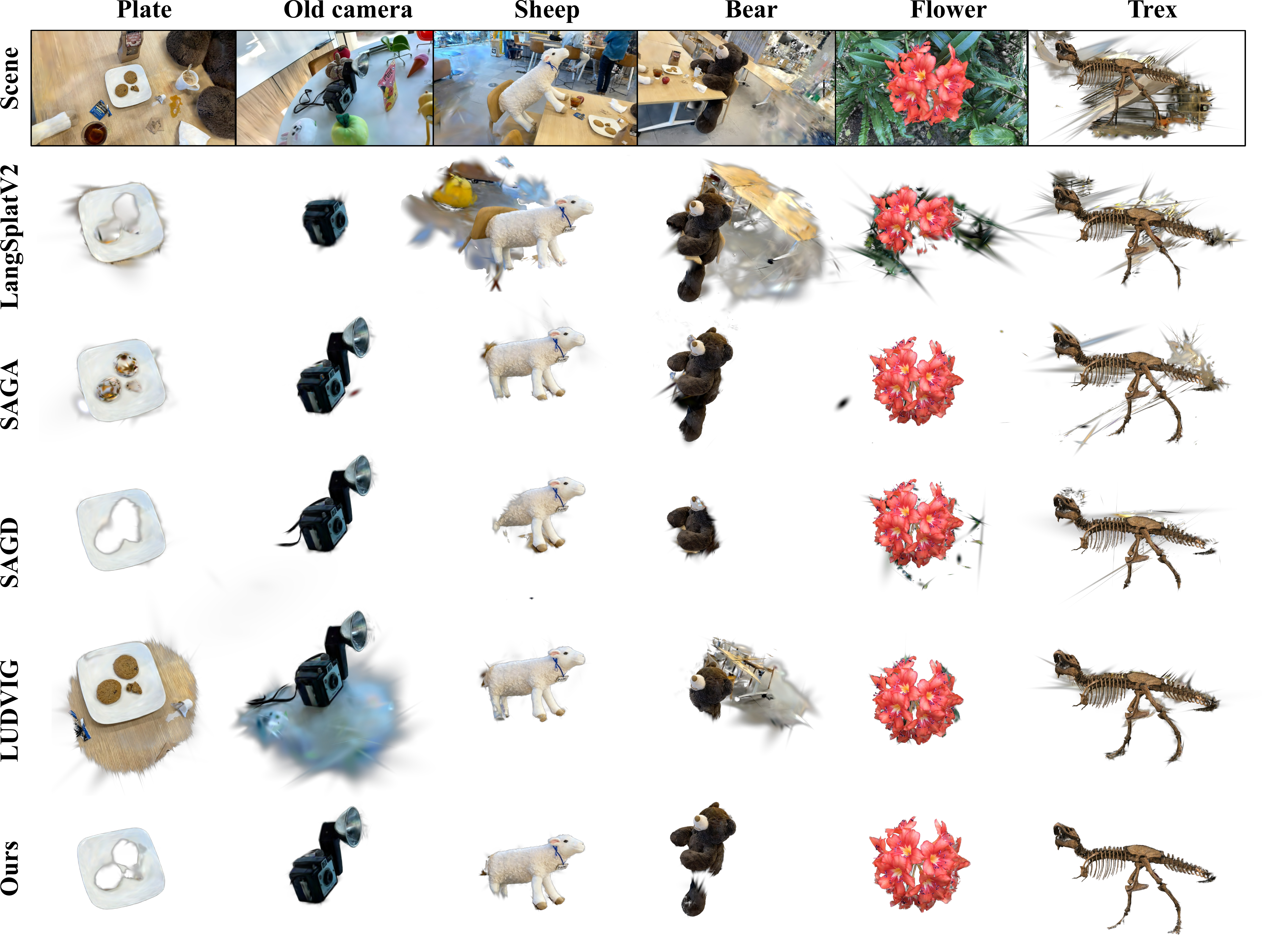}
    \captionsetup{skip=3pt}
    \caption{Qualitative comparison across diverse object scales.
        Columns are grouped by segmentation scale (small to large from left to right):
        the first four columns show objects from LERF
        and the last two from NVOS.
        VCAR produces cleaner boundaries and fewer floating artifacts
        than baseline methods across all scales.}
    \Description{Qualitative comparisons for objects of increasing scale from the LERF and NVOS
    datasets. VCAR results have cleaner object outlines and fewer detached Gaussian fragments
    than the baseline rows.}
    \label{fig:qualitative}
\end{figure*}

\subsubsection{Results on NVOS}
\cref{tab:nvos} reports the NVOS benchmark results. VCAR achieves 93.5\% mIoU and 98.6\% mAcc,
surpassing the best prior method (GaussianCut) by 1.0\% mIoU. The forward-facing capture setup
limits viewpoint diversity, making VCAR's view completeness enhancement particularly effective.

\begin{table}[t]
    \caption{Quantitative comparison of segmentation performance (mIoU(\%) and mAcc(\%) ) on
    NVOS dataset.
    The best results are highlighted in \textbf{bold} and the second-best results are
    highlighted in \underline{underlined}. Methods with $\dagger$ require per-scene training;
    without $\dagger$ are training-free.}
    \label{tab:nvos}
    \centering
    \small
    \begin{tabular}{lcc}
        \toprule
        Method                         & mIoU (\%)        & mAcc (\%)        \\
        \midrule
        SAGA$^\dagger$~\cite{SAGA}     & 90.9             & 98.3             \\
        LangSplatV2$^\dagger$~\cite{langsplatv2} & 62.2             & 88.9             \\
        COB-GS$^\dagger$~\cite{cobgs}  & 92.1             & \textbf{98.6}    \\
        \midrule
        FlashSplat~\cite{shen2024flashsplat} & 91.8             & \textbf{98.6}    \\
        GaussianCut~\cite{jain2024gaussiancut} & \underline{92.5} & 98.4             \\
        iSegMan~\cite{zhao2025isegman} & 92.0             & 98.4             \\
        SAGD~\cite{hu2024sagd}         & 90.4             & 98.2             \\
        LUDVIG~\cite{marrie2025ludvig} & 92.4             & 98.4             \\
        \textbf{VCAR (Ours)}           & \textbf{93.5}    & \textbf{98.6}    \\
        \bottomrule
    \end{tabular}
\end{table}

\subsubsection{Results on LERF}
\cref{tab:lerfovs} summarizes per-scene results on the LERF benchmark. VCAR achieves
state-of-the-art performance across all four scenes, with notable gains on \textit{kitchen}
(+15.6\% over LangSplatV2) and \textit{figurines} (+7.5\% over SAGA), where complex layouts and
severe occlusion demand comprehensive viewpoint coverage. VCAR also improves over the next best
method on \textit{ramen} (+6.1\% over SAGD) and \textit{teatime} (+6.3\% over LUDVIG),
consistently outperforming both training-based and training-free baselines.

\begin{table}[t]
    \caption{Quantitative comparison of segmentation performance (mIoU(\%) and mAcc(\%) ) on
    LERF dataset.}
    \label{tab:lerfovs}
    \centering
    \small
    \resizebox{\columnwidth}{!}{%
        \begin{tabular}{l*{4}{cc}}
            \toprule
             & \multicolumn{2}{c}{Ramen}
             & \multicolumn{2}{c}{Figurines}
             & \multicolumn{2}{c}{Teatime}
             & \multicolumn{2}{c}{Kitchen}
             \\
            \cmidrule(lr){2-3} \cmidrule(lr){4-5} \cmidrule(lr){6-7} \cmidrule(lr){8-9}
            \multirow{-2}{*}{Method}
             & mIoU                                 & mAcc             & mIoU             & mAcc
             & mIoU             & mAcc             & mIoU             & mAcc             \\
            \midrule
            Feature3DGS$^\dagger$~\cite{feature3dgs}
             & 43.7                                 & 69.8             & 40.5             & 73.4
             & 58.8             & 77.2             & 39.6             & 87.6             \\
            GaussianGrouping$^\dagger$~\cite{gaussiangrouping}
             & 45.5                                 & 68.6             & 40.0             & 74.3
             & 60.9             & 75.0             & 38.7             & 88.2             \\
            LangSplat$^\dagger$~\cite{langsplat}
             & 51.2                                 & 73.2             & 44.7             & 80.4
             & 65.1             & 88.1             & 44.5             & \underline{95.5} \\
            LangSplatV2$^\dagger$~\cite{langsplatv2}
             & 51.8                                 & 74.7             & 56.4             & 82.1
             & 72.2             & 93.2             & \underline{59.1} & 86.4             \\
            SAGA$^\dagger$~\cite{SAGA}
             & 56.4                                 & \underline{96.6} & \underline{66.3} & 98.4
             & 66.4             & 97.7             & 55.5             & 91.6             \\
            \midrule
            SAGD~\cite{hu2024sagd}
             & \underline{62.3}                     & 85.6             & 45.3             &
             \underline{99.0} & 70.9             & \underline{98.2} & 52.2             & 88.5
             \\
            LUDVIG~\cite{marrie2025ludvig}
             & 58.1                                 & 78.9             & 63.3             & 80.4
             & \underline{77.1} & 94.9             & 50.5             & 81.3             \\
            \textbf{VCAR (Ours)}
             & \textbf{68.4}                        & \textbf{98.9}    & \textbf{73.8}    &
             \textbf{99.6}    & \textbf{83.4}    & \textbf{99.4}    & \textbf{74.7}    &
             \textbf{98.1}    \\
            \bottomrule
        \end{tabular}}
\end{table}

\begin{table*}[t]
    \caption{Ablation study on NVOS dataset. "Coarse Only" gives coarse‑stage segmentation
    results; "+SSS" and "+ABR" add spherical spiral sampling or axis‑aware boundary refinement
    to the coarse baseline, respectively; "Full" combines all modules.
        }
    \label{tab:ablation}
    \centering
    \small
    \resizebox{\textwidth}{!}{%
        \begin{tabular}{l*{8}{cc}}
            \toprule
             & \multicolumn{2}{c}{Fern}
             & \multicolumn{2}{c}{Flower}
             & \multicolumn{2}{c}{Fortress}
             & \multicolumn{2}{c}{Horns}
             & \multicolumn{2}{c}{Leaves}
             & \multicolumn{2}{c}{Orchids}
             & \multicolumn{2}{c}{Trex}
             & \multicolumn{2}{c}{\textbf{Overall}}
             \\
            \cmidrule(lr){2-3} \cmidrule(lr){4-5} \cmidrule(lr){6-7}
            \cmidrule(lr){8-9} \cmidrule(lr){10-11} \cmidrule(lr){12-13} \cmidrule(lr){14-15}
            \cmidrule(lr){16-17}
            \multirow{-2}{*}{Configuration}
             & mIoU                                 & mAcc          & mIoU         & mAcc
             & mIoU & mAcc
             & mIoU                                 & mAcc          & mIoU          & mAcc
             & mIoU & mAcc & mIoU & mAcc & mIoU & mAcc \\
            \midrule
            Coarse Only
             & 82.0                                 & 94.1          & 83.2          & 95.5
             & 87.2 & 97.3
             & 92.9                                 & 98.7          & 95.2          & 99.7
             & 83.8 & 97.2 & 63.2 & 92.9 & 84.0 & 96.5 \\
            + SSS
             & 82.8                                 & 93.8          & 88.7          & 97.1
             & 95.9 & 99.2
             & 93.3                                 & 98.8          & 96.9          & 99.8
             & 88.8 & 97.6 & 73.8 & 95.7 & 88.5 & 97.4 \\
            + ABR
             & 84.8                                 & 95.0          & 93.8          & 98.5
             & 94.2 & 98.9
             & 96.1                                 & 99.3          & 95.6          & 99.7
             & 93.2 & 97.5 & 87.0 & 98.2 & 92.1 & 98.2 \\
            Full
             & \textbf{85.4}                        & \textbf{95.2} & \textbf{94.3} &
             \textbf{98.6}
             & \textbf{96.1}                        & \textbf{99.3} & \textbf{96.9} &
             \textbf{99.5}
             & \textbf{97.0}                        & \textbf{99.8} & \textbf{95.3} &
             \textbf{99.2}
             & \textbf{89.3}                        & \textbf{98.6} & \textbf{93.5} &
             \textbf{98.6}                                           \\
            \bottomrule
        \end{tabular}}
\end{table*}

\begin{table*}[t]
    \caption{Ablation study on several specific objects from LERF dataset.}
    \label{tab:ablation_lerf}
    \centering
    \small
    \resizebox{\textwidth}{!}{%
        \begin{tabular}{l*{8}{cc}}
            \toprule
             & \multicolumn{2}{c}{Figurines-Apple}
             & \multicolumn{2}{c}{Figurines-Camera}
             & \multicolumn{2}{c}{Ramen-Bowl}
             & \multicolumn{2}{c}{Ramen-Sake Cup}
             & \multicolumn{2}{c}{Teatime-Sheep}
             & \multicolumn{2}{c}{Teatime-Bear}
             & \multicolumn{2}{c}{Kitchen-Fridge}
             & \multicolumn{2}{c}{Kitchen-Toaster}
             \\
            \cmidrule(lr){2-3} \cmidrule(lr){4-5} \cmidrule(lr){6-7} \cmidrule(lr){8-9}
            \cmidrule(lr){10-11} \cmidrule(lr){12-13} \cmidrule(lr){14-15} \cmidrule(lr){16-17}
            \multirow{-2}{*}{Configuration}
             & mIoU & mAcc & mIoU & mAcc & mIoU & mAcc & mIoU & mAcc & mIoU & mAcc & mIoU & mAcc
             & mIoU & mAcc & mIoU & mAcc \\
            \midrule
            Coarse Only & 69.3 & 99.4 & 62.6 & 97.7 & 22.4 & 67.9 & 75.2 & 99.5 & 87.4 & 98.5 &
            46.3 & 77.9 & 40.7 & 85.3 & 22.9 & 95.7 \\
            + SSS       & 75.7 & 99.5 & 80.3 & 99.1 & 44.5 & 86.7 & 82.4 & 99.7 & 87.5 & 98.6 &
            69.7 & 93.7 & 85.4 & 98.3 & 74.5 & 99.5 \\
            + ABR       & 73.5 & 99.5 & 82.8 & 99.3 & 45.1 & 88.4 & 85.3 & 99.8 & 89.5 & 98.8 &
            73.5 & 94.0 & 90.7 & 98.9 & 80.5 & 99.8 \\
            Full        & \textbf{79.5} & \textbf{99.7} & \textbf{85.4} & \textbf{99.4} &
            \textbf{48.6} & \textbf{88.6} & \textbf{90.4} & \textbf{99.8} & \textbf{91.1} &
            \textbf{99.0} & \textbf{78.2} & \textbf{96.3} & \textbf{94.6} & \textbf{99.4} &
            \textbf{89.6} & \textbf{99.9} \\
            \bottomrule
        \end{tabular}}
\end{table*}

\subsubsection{Efficiency Comparison}

\begin{table}[t]
    \caption{Efficiency comparison on NVOS and LERF.
       }
    \label{tab:efficiency}
    \centering
    \small
    \resizebox{\columnwidth}{!}{%
        \begin{tabular}{l*{4}{c}}
            \toprule
                                                     & \multicolumn{2}{c}{NVOS} &
                                                     \multicolumn{2}{c}{LERF}
                                                     \\
            \cmidrule(lr){2-3} \cmidrule(lr){4-5}
            Method                                   & Training time                 & Inference
            time               & Training time    & Inference time   \\
            \midrule
            SAGA$^\dagger$~\cite{SAGA}               & $\sim$20 min             & $\sim$2 ms
            & $\sim$40 min & $\sim$2 ms   \\
            LangSplatV2$^\dagger$~\cite{langsplatv2} & $\sim$2 h                & $\sim$2 ms
            & $\sim$3 h    & $\sim$2 ms   \\
            \midrule
            LUDVIG~\cite{marrie2025ludvig}           & {0}                      & $\sim$12 min
            & {0}          & $\sim$16 min \\
            SAGD~\cite{hu2024sagd}                   & {0}                      & $\sim$2 min
            & {0}          & $\sim$3 min  \\
            \textbf{VCAR (Ours)}                     & {0}                      & $\sim$30 s
            & {0}          & $\sim$2 min  \\
            \bottomrule
        \end{tabular}}
\end{table}

\cref{tab:efficiency} compares the computational cost of VCAR with representative baselines.
Feature distillation methods require per-scene training of 20 minutes to 3 hours.
Among training-free methods, LUDVIG requires $\sim$12--16 minutes and SAGD $\sim$2--3 minutes
per object.
VCAR completes inference in $\sim$30 seconds on NVOS and $\sim$2 minutes on LERF, with the
difference primarily due to the number of images per scene.
A per-stage timing breakdown is provided in the supplementary material.
We note that training-based methods amortize their cost across multiple queries, whereas VCAR
incurs cost independently per object, making it advantageous for rapid deployment on new scenes
without re-training.
\subsection{Visualization Results}
\label{sec:qualitative}

Figure \ref{fig:qualitative} presents qualitative comparisons on selected scenes.
VCAR produces cleaner boundaries with fewer floating artifacts,
benefiting from view completeness enhancement and ABR.
In cluttered layouts (\emph{e.g.}, \textit{figurines}, \textit{kitchen}),
baseline methods include adjacent background Gaussians that produce halo-like artifacts,
which VCAR suppresses through augmented multi-view voting.
For objects with pronounced anisotropic geometry (\emph{e.g.}, \textit{orchids}),
ABR further refines boundaries by compressing only the overflowing axis
while preserving the elongated shape along well-behaved directions.

\subsection{Ablation Study}
\label{sec:ablation}

  We ablate NVOS and eight LERF objects (\cref{tab:ablation,tab:ablation_lerf}).
  According to (\cref{sec:sampling}) , most LERF objects skip SSS due to sufficient coverage
  from training views.
  Therefore, we select those that maximum angular gap exceeds $\Delta_{\text{th}}$ to trigger
  SSS for full-pipeline ablation.

\begin{figure}[t]
    \centering
    \includegraphics[width=1.0\columnwidth]{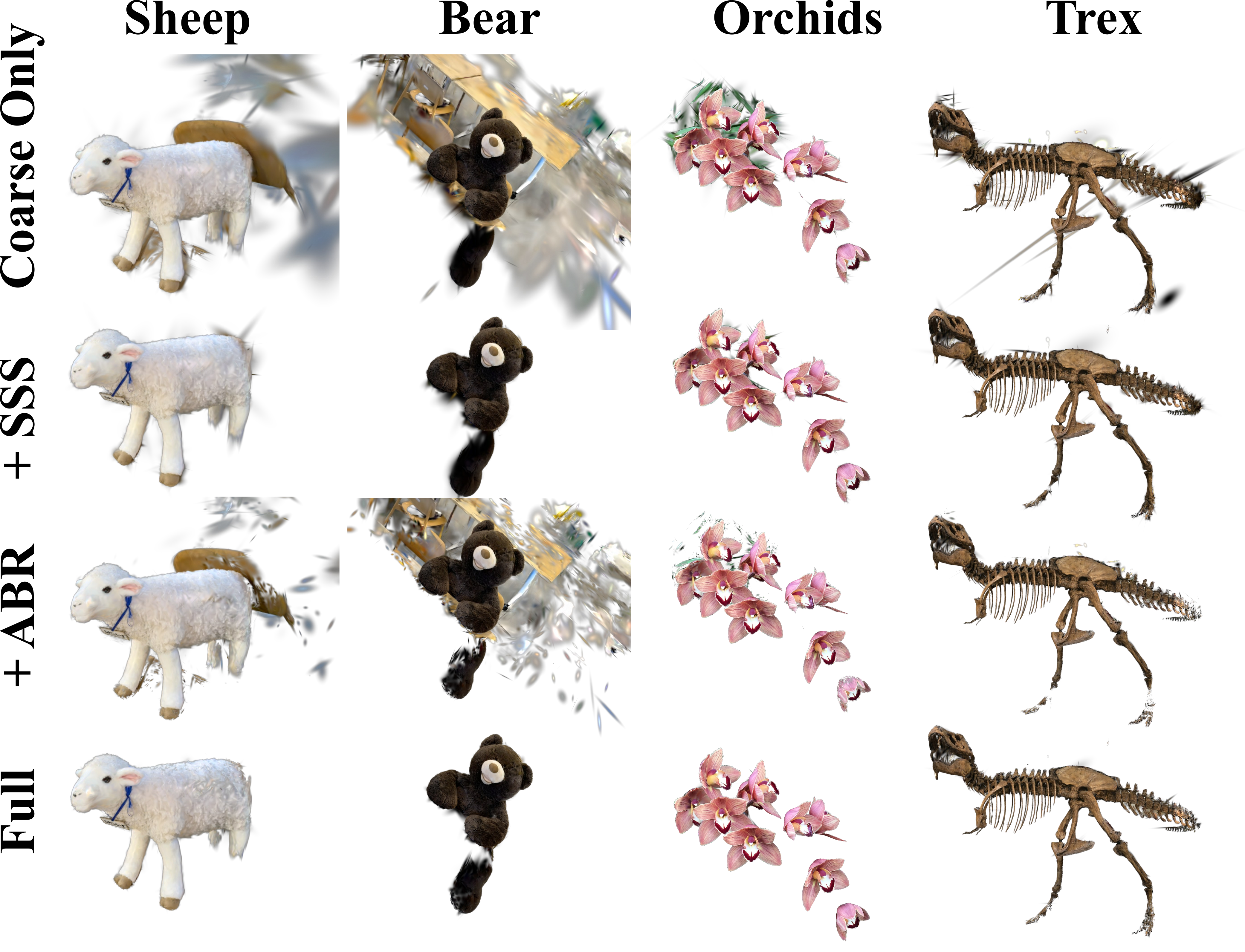}
    \caption{Visualization examples of ablation study.}
    \Description{Four NVOS objects are shown across coarse-only, spherical spiral sampling,
    boundary refinement, and full-pipeline rows, illustrating progressively cleaner and more
    complete segmentations.}
    \label{fig:ablation_visual}
\end{figure}

\noindent
  \textbf{Effect of SSS.} On NVOS, SSS raises mIoU from 84.0\% to 88.5\%, with larger gains on
  \textit{trex} (+10.6\%) and \textit{fortress} (+8.7\%) where front-facing views miss occluded
  regions.
  It also substantially improves under-sampled LERF objects such as \textit{toaster} (+51.6\%)
  and \textit{fridge} (+44.7\%).
  \textbf{Effect of ABR.} ABR raises NVOS mIoU to 92.1\%, with notable gains on \textit{trex}
  (+23.8\%) and \textit{orchids} (+9.4\%), and further improves LERF boundaries by refining only
  the dominant overflowing axis.
  \textbf{Full pipeline.} Combining both modules reaches 93.5\% mIoU on NVOS and yields
  consistent gains on the selected LERF objects.

  Figure \ref{fig:ablation_visual} illustrates the progressive refinement on representative
  scenes.
  Rather than the segmentations with noticeable boundary artifacts in the coarse stage.
  SSS reduces under-segmented regions by introducing supplementary viewpoints that increase
  angular coverage and ABR removes residual overflow for tighter boundaries.

\section{Conclusion}
\label{sec:conclusion}

We presented VCAR, a training-free coarse-to-fine segmentation framework for 3D Gaussian
Splatting that targets two important geometric contributors to blurred
boundaries: insufficient view coverage and anisotropic boundary overflow.
VCAR integrates three synergistic components: (1) object-centric spherical spiral sampling for
improved angular coverage with temporally coherent supplementary viewpoints;
(2) visibility-based weighted voting to robustly aggregate multi-view labels while excluding
invisible views; and (3) Axis-Aware Boundary Refinement (ABR), which traces 2D overflow to
specific 3D scale axes and applies targeted anisotropic compression.
Extensive experiments on NVOS and LERF show that VCAR outperforms all compared methods---both
training-based and training-free---achieving 93.5\% mIoU on NVOS and 75.1\% mIoU on LERF
(+12.8\% over the previous best), while requiring zero training and only $\sim$2 minutes of
inference per object.

\textbf{Limitations.}
The object-centric sphere assumes a roughly convex target. For highly non-convex or thin
elongated objects (e.g., cables, poles), SSS may fail to provide adequate coverage along the
object's extent.
Additionally, to correct observed 2D overflow across views,
ABR adopts a conservative minimum-scaling strategy across views.
While geometrically principled, this strict over-compression inevitably introduces minor
boundary erosion,
particularly affecting fine structures and textures.
Future work will explore adaptive, shape-aware sampling strategies for non-convex objects and
soft compression schemes that learn per-view weighting to replace the current conservative
minimum-scaling rule.

\begin{acks}
This work was supported in part by the Scientific Research Innovation Capability
Support Project for Young Faculty under Grant SRICSPYF-BS2025134; in part by the Science and
Technology Development Fund (Macau SAR) under Grant 0020/2025/RIB1; in part by the Fundamental
Research Funds for the Central Universities under Grants 21625361 and 21625360; and in part by
the Guangdong Young Science and Technology Talent Cultivation Program under Grants SKXRC2026436
and SKXRC2026434.
\end{acks}
\bibliographystyle{ACM-Reference-Format}
\balance
\bibliography{main}

@String(AAAI  = {AAAI})

@article{3DGS,
  title   = {3D Gaussian splatting for real-time radiance field rendering.},
  author  = {Kerbl, Bernhard and Kopanas, Georgios and Leimk{\"u}hler, Thomas and Drettakis, George},
  journal = {ACM Trans. Graph.},
  volume  = {42},
  number  = {4},
  pages   = {139--1},
  year    = {2023}
}

@article{Nerf,
  title     = {Nerf: Representing scenes as neural radiance fields for view synthesis},
  author    = {Mildenhall, Ben and Srinivasan, Pratul P and Tancik, Matthew and Barron, Jonathan T and Ramamoorthi, Ravi and Ng, Ren},
  journal   = {Communications of the ACM},
  volume    = {65},
  number    = {1},
  pages     = {99--106},
  year      = {2021},
  publisher = {ACM New York, NY, USA}
}

@article{3DGSApplicationSurvey,
  title   = {A survey on 3d gaussian splatting applications: Segmentation, editing, and generation},
  author  = {He, Shuting and Ji, Peilin and Yang, Yitong and Wang, Changshuo and Ji, Jiayi and Wang, Yinglin and Ding, Henghui},
  journal = {arXiv preprint arXiv:2508.09977},
  year    = {2025}
}

@article{SAM3,
  title   = {Sam 3: Segment anything with concepts},
  author  = {Carion, Nicolas and Gustafson, Laura and Hu, Yuan-Ting and Debnath, Shoubhik and Hu, Ronghang and Suris, Didac and Ryali, Chaitanya and Alwala, Kalyan Vasudev and Khedr, Haitham and Huang, Andrew and others},
  journal = {arXiv preprint arXiv:2511.16719},
  year    = {2025}
}

@article{DINOV2,
  title   = {Dinov2: Learning robust visual features without supervision},
  author  = {Oquab, Maxime and Darcet, Timoth{\'e}e and Moutakanni, Th{\'e}o and Vo, Huy and Szafraniec, Marc and Khalidov, Vasil and Fernandez, Pierre and Haziza, Daniel and Massa, Francisco and El-Nouby, Alaaeldin and others},
  journal = {arXiv preprint arXiv:2304.07193},
  year    = {2023}
}

@inproceedings{CLIP,
  title        = {Learning transferable visual models from natural language supervision},
  author       = {Radford, Alec and Kim, Jong Wook and Hallacy, Chris and Ramesh, Aditya and Goh, Gabriel and Agarwal, Sandhini and Sastry, Girish and Askell, Amanda and Mishkin, Pamela and Clark, Jack and others},
  booktitle    = {International conference on machine learning},
  pages        = {8748--8763},
  year         = {2021},
  organization = {PmLR}
}

@inproceedings{gaussianpro,
  title     = {Gaussianpro: 3d gaussian splatting with progressive propagation},
  author    = {Cheng, Kai and Long, Xiaoxiao and Yang, Kaizhi and Yao, Yao and Yin, Wei and Ma, Yuexin and Wang, Wenping and Chen, Xuejin},
  booktitle = {Forty-first International Conference on Machine Learning},
  year      = {2024}
}

@inproceedings{Scaffold,
  title     = {Scaffold-gs: Structured 3d gaussians for view-adaptive rendering},
  author    = {Lu, Tao and Yu, Mulin and Xu, Linning and Xiangli, Yuanbo and Wang, Limin and Lin, Dahua and Dai, Bo},
  booktitle = {Proceedings of the IEEE/CVF conference on computer vision and pattern recognition},
  pages     = {20654--20664},
  year      = {2024}
}

@inproceedings{mipsplatting,
  title     = {Mip-splatting: Alias-free 3d gaussian splatting},
  author    = {Yu, Zehao and Chen, Anpei and Huang, Binbin and Sattler, Torsten and Geiger, Andreas},
  booktitle = {Proceedings of the IEEE/CVF conference on computer vision and pattern recognition},
  pages     = {19447--19456},
  year      = {2024}
}

@article{lightgaussian,
  title   = {Lightgaussian: Unbounded 3d gaussian compression with 15x reduction and 200+ fps},
  author  = {Fan, Zhiwen and Wang, Kevin and Wen, Kairun and Zhu, Zehao and Xu, Dejia and Wang, Zhangyang},
  journal = {Advances in neural information processing systems},
  volume  = {37},
  pages   = {140138--140158},
  year    = {2024}
}

@inproceedings{compact3dgs,
  title     = {Compact 3d gaussian representation for radiance field},
  author    = {Lee, Joo Chan and Rho, Daniel and Sun, Xiangyu and Ko, Jong Hwan and Park, Eunbyung},
  booktitle = {Proceedings of the IEEE/CVF Conference on Computer Vision and Pattern Recognition},
  pages     = {21719--21728},
  year      = {2024}
}

@inproceedings{gaufre,
  title        = {Gaufre: Gaussian deformation fields for real-time dynamic novel view synthesis},
  author       = {Liang, Yiqing and Khan, Numair and Li, Zhengqin and Nguyen-Phuoc, Thu and Lanman, Douglas and Tompkin, James and Xiao, Lei},
  booktitle    = {2025 IEEE/CVF Winter Conference on Applications of Computer Vision (WACV)},
  pages        = {2642--2652},
  year         = {2025},
  organization = {IEEE}
}

@inproceedings{4dgs,
  title     = {4d gaussian splatting for real-time dynamic scene rendering},
  author    = {Wu, Guanjun and Yi, Taoran and Fang, Jiemin and Xie, Lingxi and Zhang, Xiaopeng and Wei, Wei and Liu, Wenyu and Tian, Qi and Wang, Xinggang},
  booktitle = {Proceedings of the IEEE/CVF conference on computer vision and pattern recognition},
  pages     = {20310--20320},
  year      = {2024}
}

@inproceedings{pixelsplat,
  title     = {pixelsplat: 3d gaussian splats from image pairs for scalable generalizable 3d reconstruction},
  author    = {Charatan, David and Li, Sizhe Lester and Tagliasacchi, Andrea and Sitzmann, Vincent},
  booktitle = {Proceedings of the IEEE/CVF conference on computer vision and pattern recognition},
  pages     = {19457--19467},
  year      = {2024}
}

@inproceedings{fsgs,
  title        = {Fsgs: Real-time few-shot view synthesis using gaussian splatting},
  author       = {Zhu, Zehao and Fan, Zhiwen and Jiang, Yifan and Wang, Zhangyang},
  booktitle    = {European conference on computer vision},
  pages        = {145--163},
  year         = {2024},
  organization = {Springer}
}

@inproceedings{Gaussianeditor,
  title     = {Gaussianeditor: Swift and controllable 3d editing with gaussian splatting},
  author    = {Chen, Yiwen and Chen, Zilong and Zhang, Chi and Wang, Feng and Yang, Xiaofeng and Wang, Yikai and Cai, Zhongang and Yang, Lei and Liu, Huaping and Lin, Guosheng},
  booktitle = {Proceedings of the IEEE/CVF conference on computer vision and pattern recognition},
  pages     = {21476--21485},
  year      = {2024}
}

@inproceedings{qiu2024language,
  title        = {Language-driven physics-based scene synthesis and editing via feature splatting},
  author       = {Qiu, Ri-Zhao and Yang, Ge and Zeng, Weijia and Wang, Xiaolong},
  booktitle    = {European conference on computer vision},
  pages        = {368--383},
  year         = {2024},
  organization = {Springer}
}

@inproceedings{physgaussian,
  title     = {Physgaussian: Physics-integrated 3d gaussians for generative dynamics},
  author    = {Xie, Tianyi and Zong, Zeshun and Qiu, Yuxing and Li, Xuan and Feng, Yutao and Yang, Yin and Jiang, Chenfanfu},
  booktitle = {Proceedings of the IEEE/CVF Conference on Computer Vision and Pattern Recognition},
  pages     = {4389--4398},
  year      = {2024}
}

@inproceedings{luciddreamer,
  title     = {Luciddreamer: Towards high-fidelity text-to-3d generation via interval score matching},
  author    = {Liang, Yixun and Yang, Xin and Lin, Jiantao and Li, Haodong and Xu, Xiaogang and Chen, Yingcong},
  booktitle = {Proceedings of the IEEE/CVF conference on computer vision and pattern recognition},
  pages     = {6517--6526},
  year      = {2024}
}

@article{dreamgaussian,
  title   = {Dreamgaussian: Generative gaussian splatting for efficient 3d content creation},
  author  = {Tang, Jiaxiang and Ren, Jiawei and Zhou, Hang and Liu, Ziwei and Zeng, Gang},
  journal = {arXiv preprint arXiv:2309.16653},
  year    = {2023}
}

@inproceedings{langsplat,
  title     = {Langsplat: 3d language gaussian splatting},
  author    = {Qin, Minghan and Li, Wanhua and Zhou, Jiawei and Wang, Haoqian and Pfister, Hanspeter},
  booktitle = {Proceedings of the IEEE/CVF Conference on Computer Vision and Pattern Recognition},
  pages     = {20051--20060},
  year      = {2024}
}

@article{langsplatv2,
  title   = {Langsplatv2: High-dimensional 3d language gaussian splatting with 450+ fps},
  author  = {Li, Wanhua and Zhao, Yujie and Qin, Minghan and Liu, Yang and Cai, Yuanhao and Gan, Chuang and Pfister, Hanspeter},
  journal = {arXiv preprint arXiv:2507.07136},
  year    = {2025}
}

@inproceedings{feature3dgs,
  title     = {Feature 3dgs: Supercharging 3d gaussian splatting to enable distilled feature fields},
  author    = {Zhou, Shijie and Chang, Haoran and Jiang, Sicheng and Fan, Zhiwen and Zhu, Zehao and Xu, Dejia and Chari, Pradyumna and You, Suya and Wang, Zhangyang and Kadambi, Achuta},
  booktitle = {Proceedings of the IEEE/CVF Conference on Computer Vision and Pattern Recognition},
  pages     = {21676--21685},
  year      = {2024}
}

@inproceedings{shi2024language,
  title     = {Language embedded 3d gaussians for open-vocabulary scene understanding},
  author    = {Shi, Jin-Chuan and Wang, Miao and Duan, Hao-Bin and Guan, Shao-Hua},
  booktitle = {Proceedings of the IEEE/CVF Conference on Computer Vision and Pattern Recognition},
  pages     = {5333--5343},
  year      = {2024}
}

@article{gao2024fast,
  title   = {Fast and efficient: Mask neural fields for 3d scene segmentation},
  author  = {Gao, Zihan and Li, Lingling and Jiao, Licheng and Liu, Fang and Liu, Xu and Ma, Wenping and Guo, Yuwei and Yang, Shuyuan},
  journal = {arXiv preprint arXiv:2407.01220},
  year    = {2024}
}

@article{lseg,
  title   = {Language-driven semantic segmentation},
  author  = {Li, Boyi and Weinberger, Kilian Q and Belongie, Serge and Koltun, Vladlen and Ranftl, Ren{\'e}},
  journal = {arXiv preprint arXiv:2201.03546},
  year    = {2022}
}

@article{li2024langsurf,
  title   = {Langsurf: Language-embedded surface gaussians for 3d scene understanding},
  author  = {Li, Hao and Qin, Roy and Zou, Zhengyu and He, Diqi and Li, Bohan and Dai, Bingquan and Zhang, Dingewn and Han, Junwei},
  journal = {arXiv preprint arXiv:2412.17635},
  year    = {2024}
}

@article{wu2024opengaussian,
  title   = {Opengaussian: Towards point-level 3d gaussian-based open vocabulary understanding},
  author  = {Wu, Yanmin and Meng, Jiarui and Li, Haijie and Wu, Chenming and Shi, Yahao and Cheng, Xinhua and Zhao, Chen and Feng, Haocheng and Ding, Errui and Wang, Jingdong and others},
  journal = {Advances in Neural Information Processing Systems},
  volume  = {37},
  pages   = {19114--19138},
  year    = {2024}
}

@article{zuo2025fmgs,
  title     = {Fmgs: Foundation model embedded 3d gaussian splatting for holistic 3d scene understanding},
  author    = {Zuo, Xingxing and Samangouei, Pouya and Zhou, Yunwen and Di, Yan and Li, Mingyang},
  journal   = {International Journal of Computer Vision},
  volume    = {133},
  number    = {2},
  pages     = {611--627},
  year      = {2025},
  publisher = {Springer}
}

@inproceedings{kirillov2023segment,
  title     = {Segment anything},
  author    = {Kirillov, Alexander and Mintun, Eric and Ravi, Nikhila and Mao, Hanzi and Rolland, Chloe and Gustafson, Laura and Xiao, Tete and Whitehead, Spencer and Berg, Alexander C and Lo, Wan-Yen and others},
  booktitle = {Proceedings of the IEEE/CVF international conference on computer vision},
  pages     = {4015--4026},
  year      = {2023}
}

@inproceedings{kerr2023lerf,
  title     = {Lerf: Language embedded radiance fields},
  author    = {Kerr, Justin and Kim, Chung Min and Goldberg, Ken and Kanazawa, Angjoo and Tancik, Matthew},
  booktitle = {Proceedings of the IEEE/CVF international conference on computer vision},
  pages     = {19729--19739},
  year      = {2023}
}

@inproceedings{gaussiangrouping,
  title        = {Gaussian grouping: Segment and edit anything in 3d scenes},
  author       = {Ye, Mingqiao and Danelljan, Martin and Yu, Fisher and Ke, Lei},
  booktitle    = {European conference on computer vision},
  pages        = {162--179},
  year         = {2024},
  organization = {Springer}
}

@article{hu2024sagd,
  title   = {SAGD: Boundary-enhanced segment anything in 3D Gaussian via Gaussian decomposition},
  author  = {Hu, Xu and Wang, Yuxi and Fan, Lue and Luo, Chuanchen and Fan, Junsong and Lei, Zhen and Li, Qing and Peng, Junran and Zhang, Zhaoxiang},
  journal = {arXiv preprint arXiv:2401.17857},
  year    = {2024}
}

@article{lyu2024gaga,
  title   = {Gaga: Group any gaussians via 3d-aware memory bank},
  author  = {Lyu, Weijie and Li, Xueting and Kundu, Abhijit and Tsai, Yi-Hsuan and Yang, Ming-Hsuan},
  journal = {arXiv preprint arXiv:2404.07977},
  year    = {2024}
}

@article{jain2024gaussiancut,
  title   = {Gaussiancut: Interactive segmentation via graph cut for 3d gaussian splatting},
  author  = {Jain, Umangi and Mirzaei, Ashkan and Gilitschenski, Igor},
  journal = {Advances in Neural Information Processing Systems},
  volume  = {37},
  pages   = {89184--89212},
  year    = {2024}
}

@inproceedings{shen2024flashsplat,
  title        = {Flashsplat: 2d to 3d gaussian splatting segmentation solved optimally},
  author       = {Shen, Qiuhong and Yang, Xingyi and Wang, Xinchao},
  booktitle    = {European Conference on Computer Vision},
  pages        = {456--472},
  year         = {2024},
  organization = {Springer}
}

@inproceedings{ying2024omniseg3d,
  title     = {Omniseg3d: Omniversal 3d segmentation via hierarchical contrastive learning},
  author    = {Ying, Haiyang and Yin, Yixuan and Zhang, Jinzhi and Wang, Fan and Yu, Tao and Huang, Ruqi and Fang, Lu},
  booktitle = {Proceedings of the IEEE/CVF Conference on Computer Vision and Pattern Recognition},
  pages     = {20612--20622},
  year      = {2024}
}

@inproceedings{SAGA,
  title     = {Segment any 3d gaussians},
  author    = {Cen, Jiazhong and Fang, Jiemin and Yang, Chen and Xie, Lingxi and Zhang, Xiaopeng and Shen, Wei and Tian, Qi},
  booktitle = {Proceedings of the AAAI conference on artificial intelligence},
  volume    = {39},
  number    = {2},
  pages     = {1971--1979},
  year      = {2025}
}

@inproceedings{clickgaussian,
  title        = {Click-gaussian: Interactive segmentation to any 3d gaussians},
  author       = {Choi, Seokhun and Song, Hyeonseop and Kim, Jaechul and Kim, Taehyeong and Do, Hoseok},
  booktitle    = {European Conference on Computer Vision},
  pages        = {289--305},
  year         = {2024},
  organization = {Springer}
}

@inproceedings{zhao2025isegman,
  title     = {isegman: Interactive segment-and-manipulate 3d gaussians},
  author    = {Zhao, Yian and Xu, Wanshi and Zheng, Ruochong and Qiao, Pengchong and Liu, Chang and Chen, Jie},
  booktitle = {Proceedings of the Computer Vision and Pattern Recognition Conference},
  pages     = {661--670},
  year      = {2025}
}

@inproceedings{marrie2025ludvig,
  title     = {Ludvig: Learning-free uplifting of 2d visual features to gaussian splatting scenes},
  author    = {Marrie, Juliette and M{\'e}n{\'e}gaux, Romain and Arbel, Michael and Larlus, Diane and Mairal, Julien},
  booktitle = {Proceedings of the IEEE/CVF International Conference on Computer Vision},
  pages     = {7440--7450},
  year      = {2025}
}

@inproceedings{cobgs,
  title     = {Cob-gs: Clear object boundaries in 3dgs segmentation based on boundary-adaptive gaussian splitting},
  author    = {Zhang, Jiaxin and Jiang, Junjun and Chen, Youyu and Jiang, Kui and Liu, Xianming},
  booktitle = {Proceedings of the IEEE/CVF Conference on Computer Vision and Pattern Recognition},
  pages     = {19335--19344},
  year      = {2025}
}

@inproceedings{ren2022nvos,
  title     = {Neural volumetric object selection},
  author    = {Ren, Zhongzheng and Agarwala, Aseem and Russell, Bryan and Schwing, Alexander G and Wang, Oliver},
  booktitle = {Proceedings of the IEEE/CVF Conference on Computer Vision and Pattern Recognition},
  pages     = {6133--6142},
  year      = {2022}
}

@article{paszke2019pytorch,
  title   = {Pytorch: An imperative style, high-performance deep learning library},
  author  = {Paszke, Adam and Gross, Sam and Massa, Francisco and Lerer, Adam and Bradbury, James and Chanan, Gregory and Killeen, Trevor and Lin, Zeming and Gimelshein, Natalia and Antiga, Luca and others},
  journal = {Advances in neural information processing systems},
  volume  = {32},
  year    = {2019}
}

@article{ye2025gsplat,
  title   = {gsplat: An open-source library for Gaussian splatting},
  author  = {Ye, Vickie and Li, Ruilong and Kerr, Justin and Turkulainen, Matias and Yi, Brent and Pan, Zhuoyang and Seiskari, Otto and Ye, Jianbo and Hu, Jeffrey and Tancik, Matthew and others},
  journal = {Journal of Machine Learning Research},
  volume  = {26},
  number  = {34},
  pages   = {1--17},
  year    = {2025}
}

@inproceedings{zhang2025labelgs,
  title        = {LabelGS: Label-Aware 3D Gaussian Splatting for 3D Scene Segmentation},
  author       = {Zhang, Yupeng and Zheng, Dezhi and Lu, Ping and Zhang, Han and Wang, Lei and Xiang, Liping and Luo, Cheng and Deng, Kaijun and Fu, Xiaowen and Shen, Linlin and others},
  booktitle    = {Chinese Conference on Pattern Recognition and Computer Vision (PRCV)},
  pages        = {47--61},
  year         = {2025},
  organization = {Springer}
}

@inproceedings{chacko2025lifting,
  title        = {Lifting by gaussians: A simple, fast and flexible method for 3d instance segmentation},
  author       = {Chacko, Rohan and H{\"a}ni, Nicolai and Khaliullin, Eldar and Sun, Lin and Lee, Douglas},
  booktitle    = {2025 IEEE/CVF Winter Conference on Applications of Computer Vision (WACV)},
  pages        = {3497--3507},
  year         = {2025},
  organization = {IEEE}
}

@article{liao2025clipgs,
  title     = {CLIP-GS: CLIP-Informed Gaussian Splatting for View-Consistent 3D Indoor Semantic Understanding},
  author    = {Liao, Guibiao and Li, Jiankun and Bao, Zhenyu and Ye, Xiaoqing and Li, Qing and Liu, Kanglin},
  journal   = {ACM Transactions on Multimedia Computing, Communications and Applications},
  volume    = {21},
  number    = {8},
  pages     = {1--24},
  year      = {2025},
  publisher = {ACM New York, NY}
}

@article{liao2026gaussiantrimmer,
  title   = {GaussianTrimmer: Online Trimming Boundaries for 3DGS Segmentation},
  author  = {Liao, Liwei and Wang, Ronggang},
  journal = {arXiv preprint arXiv:2601.12683},
  year    = {2026}
}

@inproceedings{yuan2025robust,
  title={Robust and Efficient 3D Gaussian Splatting for Urban Scene Reconstruction},
  author={Yuan, Zhensheng and Huang, Haozhi and Xiong, Zhen and Wang, Di and Yang, Guanghua},
  booktitle={Proceedings of the IEEE/CVF International Conference on Computer Vision},
  pages={26209--26219},
  year={2025}
}

@inproceedings{chen2025slgaussian,
  title={Slgaussian: Fast language gaussian splatting in sparse views},
  author={Chen, Kangjie and Dai, BingQuan and Qin, Minghan and Zhang, Dongbin and Li, Peihao and Zou, Yingshuang and Wang, Haoqian},
  booktitle={Proceedings of the 33rd ACM International Conference on Multimedia},
  pages={3047--3056},
  year={2025}
}

@inproceedings{bao2025segwild,
  title={Seg-wild: Interactive segmentation based on 3d gaussian splatting for unconstrained image collections},
  author={Bao, Yongtang and Tang, Chengjie and Wang, Yuze and Li, Haojie},
  booktitle={Proceedings of the 33rd ACM International Conference on Multimedia},
  pages={8567--8576},
  year={2025}
}

@inproceedings{li2024fsh3d,
  title={Fsh3d: 3d representation via fibonacci spherical harmonics},
  author={Li, Zikuan and Huang, Anyi and Jia, Wenru and Wu, Qiaoyun and Wei, Mingqiang and Wang, Jun},
  booktitle={Computer Graphics Forum},
  volume={43},
  number={7},
  pages={e15231},
  year={2024},
  organization={Wiley Online Library}
}

@article{ravi2024sam,
  title={Sam 2: Segment anything in images and videos},
  author={Ravi, Nikhila and Gabeur, Valentin and Hu, Yuan-Ting and Hu, Ronghang and Ryali, Chaitanya and Ma, Tengyu and Khedr, Haitham and R{\"a}dle, Roman and Rolland, Chloe and Gustafson, Laura and others},
  journal={arXiv preprint arXiv:2408.00714},
  year={2024}
}

\clearpage
\nobalance

\setlength{\dbltextfloatsep}{8pt plus 2pt minus 2pt}
\setcounter{section}{0}
\setcounter{subsection}{0}
\setcounter{equation}{0}
\setcounter{table}{0}
\setcounter{figure}{0}
\renewcommand{\thesection}{\Alph{section}}
\renewcommand{\thesubsection}{\thesection.\arabic{subsection}}
\renewcommand{\theequation}{\thesection.\arabic{equation}}
\renewcommand{\thetable}{\thesection.\arabic{table}}
\renewcommand{\thefigure}{\thesection.\arabic{figure}}
\counterwithin{equation}{section}
\counterwithin{table}{section}
\counterwithin{figure}{section}

\twocolumn[
  \begin{center}
    {\LARGE\bfseries VCAR: Training-Free 3DGS Segmentation via View Completeness\par}
    {\LARGE\bfseries and Axis-Aware Boundary Refinement\par}
    \vspace{0.45em}
    {\Large\bfseries Supplementary Material\par}
    \vspace{0.9em}
  \end{center}
]

\section{View Completeness Assessment: Detailed Workflow and Rationale}
\label{sec:supp_camera_filtering}

This section expands on two implementation details omitted from Section~3.3.2 of the main paper due to space constraints: (1) \cref{sec:supp_filtering}: the camera validity filtering criterion; (2) \cref{sec:supp_probing}: the Fibonacci-lattice probing mechanism used to evaluate
$\Delta_{\max}$.
We also include a brief rationale for this probing strategy
and the geometric interpretation of the $90^\circ$ trigger threshold.
Unless otherwise specified, notation in this section follows the main paper. The key inherited symbols used in this section are:
\begin{itemize}
  \item $\mathcal{V}^{\text{train}}=\{v^{(1)}, \ldots, v^{(M)}\}$ denotes the viewpoints from the training set (Section~3.1).
  \item $\mathcal{G}^{\text{coarse}}$ denotes the coarse segmentation result (Section~3.2).
  \item $\mathcal{P}=\{\boldsymbol{\mu}_i \mid g_i \in \mathcal{G}^{\text{coarse}}\}$ denotes the coarse foreground Gaussian positions (Section~3.3.1).
  \item $\boldsymbol{c}$ denotes the robust object center of the object-centric sphere (Section~3.3.1).
  \item $v^{(j)}_{\text{pos}}$ and $\mathbf{h}_j$ denote the $j$-th camera position and its unit heading toward $\boldsymbol{c}$, respectively(main paper, Section~3.3.2).
  \item $\{\mathbf{t}_i\}_{i=1}^{K}$ and $\Delta_{\max}$ denote the $K$ Fibonacci probing directions and the maximum angular gap, respectively(main paper, Section~3.3.2).
\end{itemize}

\subsection{Camera Validity Filtering}
\label{sec:supp_filtering}


Since not every camera view in the training set can effectively observe the target object, we filter out those invalid camera views before computing the view coverage.

Let $N_{\text{c}}$
denote the number of coarse foreground Gaussians $|\mathcal{P}|$.
For each training camera $v^{(j)}$, we test every $\boldsymbol{\mu}_i \in \mathcal{P}$ for visibility using the same depth-and-bounds criterion of Section~3.2 (Eq.~1 of the main paper).
Let the boolean condition $\mathrm{visible}_i^{(j)}$ denote whether $\boldsymbol{\mu}_i$ is successfully projected into the valid image bounds of $v^{(j)}$ with a positive depth.
We then compute the visible foreground ratio:
\begin{equation}
  \rho^{(j)} = \frac{1}{N_{\text{c}}}
  \sum_{i=1}^{N_{\text{c}}}
  \mathbb{1}\bigl[\mathrm{visible}_i^{(j)}\bigr]
\end{equation}

We adopt $\rho_{\min} = 0.2$ (20\%) as the default validity threshold. If the threshold is too low, cameras that observe only a negligible portion of the target would be retained and would falsely mark their directions as covered. Conversely, if the threshold is too high, cameras that still provide meaningful partial observations, such as side views in forward-facing captures, would be discarded, making the assessment overly pessimistic.

Accordingly, the set of valid cameras is denoted
$\mathcal{V}^{\text{valid}}
  = \{v^{(j)} \in \mathcal{V}^{\text{train}}
  \mid \rho^{(j)} \geq \rho_{\min}\}$,
and the corresponding unit viewpoint directions are
\begin{equation}
  \label{eq:heading}
  \mathbf{h}_j
  = \frac{v^{(j)}_{\text{pos}} - \boldsymbol{c}}
  {\|v^{(j)}_{\text{pos}} - \boldsymbol{c}\|_2}
\end{equation}
where $v^{(j)}_{\text{pos}}$ is the camera position of $v^{(j)}$.

\subsection{Fibonacci-Lattice Probing}
\label{sec:supp_probing}

The maximum angular gap $\Delta_{\max}$
(Eq.~6 of the main paper) is computed by scattering $K=2000$ test directions
on the unit sphere via a Fibonacci lattice \cite{li2024fsh3d} and measuring the worst-case angular distance
to the nearest valid camera.
Below we detail the lattice construction and the efficient matrix computation.

\subsubsection{Fibonacci lattice construction.}
For the $i$-th test direction ($i = 0, \ldots, K-1$),
the polar and azimuthal angles are
\begin{equation}
  \varphi_i = \arccos \bigl(1 - 2(i + 0.5)/K\bigr),
  \qquad
  \vartheta_i = \pi(1 + \sqrt{5})\,i
\end{equation}
where $\pi(1+\sqrt{5}) \approx 137.508^\circ$ is the golden angle.
The term $1 - 2(i+0.5)/K$
distributes $\cos\varphi$ uniformly in $[-1,1]$,
producing equal-area latitude slices and avoiding the polar clustering of regular latitude--longitude grids.
Successive golden-angle rotations in azimuth place each new point into a previously under-sampled region
of its latitude band, yielding a near-uniform distribution over the sphere in $O(K)$ time.
The Cartesian coordinates are
\begin{equation}
  \mathbf{t}_i
  = \bigl(
  \sin\varphi_i\cos\vartheta_i,\;
  \sin\varphi_i\sin\vartheta_i,\;
  \cos\varphi_i
  \bigr)^\top
\end{equation}

\subsubsection{Coverage computation.}
We form the cosine similarity matrix
\begin{equation}
  \mathbf{C} \in \mathbb{R}^{K \times |\mathcal{V}^{\text{valid}}|},
  \qquad
  C_{ij} = \mathbf{t}_i^\top \mathbf{h}_j,
\end{equation}
and obtain the per-probe angular distance and maximum angular gap to identify the most under-observed direction on the sphere:
\begin{equation}
  \delta_i = \arccos\!\left(\max\limits_{j} C_{ij}\right), \qquad
  \Delta_{\max} = \max\limits_{i} \delta_i .
\end{equation}

Compared with pairwise camera-angle statistics,
Fibonacci lattice probing directly measures the target quantity in view completeness:
the worst-case angular distance from any spherical direction to its nearest valid camera.
With $K=2000$, this estimate is obtained efficiently by one $\mathbf{T}\mathbf{H}^\top$ matrix product followed by row-wise and global $\max$ reductions, taking only 2--3 seconds in scenes with hundreds of cameras.
Moreover, $\Delta_{\text{th}}=90^\circ$ has a clear geometric meaning: it corresponds to a hemisphere-scale blind spot (solid angle $2\pi$), making it a principled trigger for SSS.

\section{Axis-Aware Boundary Refinement: Detailed Derivation}
\label{sec:supp_abr}

Section~3.4 of the main paper introduces Axis-Aware Boundary Refinement (ABR)
and gives its principal covariance decomposition and compression rule.
This section expands the intermediate steps for projected overflow detection,
dominant-axis attribution, directional compression, and multi-view
reconciliation.
We follow the notation of the main paper. Throughout the single-view derivation below,
we fix a view $v^{(j)}$ and omit $(j)$ from all view-conditioned quantities,
e.g.,
$\boldsymbol{m}_i^{(j)}\mapsto\boldsymbol{m}_i$,
$\boldsymbol{\Sigma}_{2D,i}^{(j)}\mapsto
\boldsymbol{\Sigma}_{2D,i}$, and
$\mathcal{M}^{(j)}\mapsto\mathcal{M}$. We restore the view index only when
observations are collected across views, as in $o_i^{(j)}$ and
$f_{i,d}^{(j,\boldsymbol{e})}$. The 3D primitive $g_i$ itself is
view-independent and therefore never carries a view superscript.

\subsection{Projected Ellipse and Overflow Detection}
\label{sec:supp_abr_detection}

For a visible Gaussian $g_i$, let $\boldsymbol{m}_i$ and
$\boldsymbol{\Sigma}_{2D,i}$ denote its projected center and 2D covariance.
The eigendecomposition of the covariance is
\begin{equation}
  \boldsymbol{\Sigma}_{2D,i}
  =
  \mathbf{U}_i
  \operatorname{diag}(\lambda_1,\lambda_2)
  \mathbf{U}_i^\top,
  \qquad
  \lambda_1 \geq \lambda_2,
\end{equation}
where the columns $\vec{e}_1$ and $\vec{e}_2$ of $\mathbf{U}_i$ are the
principal directions of the projected ellipse. With the diagnostic cutoff
$\sigma_c=3$ adopted by ABR and used in the main paper, its semi-axis lengths
are
$a_i=\sigma_c\sqrt{\lambda_1}$ and
$b_i=\sigma_c\sqrt{\lambda_2}$. The four diagnostic endpoints are therefore
\begin{equation}
  \label{eq:supp_ellipse_endpoints}
  \mathcal{E}_i
  =
  \left\{
    \boldsymbol{m}_i \pm a_i\vec{e}_1,\;
    \boldsymbol{m}_i \pm b_i\vec{e}_2
  \right\}.
\end{equation}

Let $\mathcal{M}$ and $\Omega$ denote the binary foreground mask and image
domain of the fixed view. An observation is eligible for ABR only when the
projected center lies in $\Omega$ and on the foreground of $\mathcal{M}$.
For an eligible observation, we record an overflow whenever at least one
endpoint lies outside the image or maps to the background:
\begin{equation}
  o_i
  =
  \mathbb{1}\!\left[
    \exists\,\boldsymbol{e}\in\mathcal{E}_i
    :
    \boldsymbol{e}\notin\Omega
    \;\lor\;
    \mathcal{M}(\operatorname{round}(\boldsymbol{e}))=0
  \right].
\end{equation}

Restoring the view index, this indicator becomes $o_i^{(j)}$ and contributes
to the multi-view count $n_i^{\mathrm{ovf}}$ used below.
This endpoint test is directional: an elongated Gaussian is not penalized
merely for having a large major axis if both endpoints remain within the
foreground support. Importantly, $\vec{e}_1$ and $\vec{e}_2$ are not direct
projections of two individual 3D local axes. Each projected eigenvector
depends jointly on all three axes through the covariance projection, which
motivates the attribution below.

\subsection{Dominant 3D Axis Attribution}
\label{sec:supp_abr_attribution}

Let $\mathbf{R}_i=[\mathbf{r}_1,\mathbf{r}_2,\mathbf{r}_3]$ and
$\mathbf{s}_i=(s_1,s_2,s_3)$ denote the rotation and scale of $g_i$.
Its 3D covariance can be expanded by local axis:
\begin{equation}
  \boldsymbol{\Sigma}_i
  =
  \mathbf{R}_i
  \operatorname{diag}(s_1^2,s_2^2,s_3^2)
  \mathbf{R}_i^\top
  =
  \sum_{d=1}^{3}s_d^2\mathbf{r}_d\mathbf{r}_d^\top.
\end{equation}
For the fixed view, let $\mathbf{W}_{R}$ be the rotation block of the
world-to-camera transformation and let $\mathbf{J}_i$ be the perspective
Jacobian evaluated at the center of $g_i$. Their product
$\mathbf{M}_i=\mathbf{J}_i\mathbf{W}_{R}$ maps a small world-space
displacement to the image plane. Substituting the local-axis expansion
above gives the scale-dependent geometric projection covariance
\begin{align}
  \boldsymbol{\Sigma}_{2D,i}
  &=
  \mathbf{M}_i\boldsymbol{\Sigma}_i\mathbf{M}_i^\top \notag\\
  &=
  \mathbf{M}_i
  \left(
    \sum_{d=1}^{3}s_d^2\mathbf{r}_d\mathbf{r}_d^\top
  \right)
  \mathbf{M}_i^\top \notag\\
  &=
  \sum_{d=1}^{3}
  s_d^2
  \bigl(\mathbf{M}_i\mathbf{r}_d\bigr)
  \bigl(\mathbf{M}_i\mathbf{r}_d\bigr)^\top \notag\\
  &=
  \sum_{d=1}^{3}
  s_d^2\mathbf{q}_d\mathbf{q}_d^\top,
  \qquad
  \mathbf{q}_d=\mathbf{M}_i\mathbf{r}_d.
  \label{eq:supp_cov_decomp}
\end{align}

Consider one overflowing endpoint, with unit direction
$\mathbf{u}\in\{\pm\vec{e}_1,\pm\vec{e}_2\}$ and the corresponding
eigenvalue $\lambda\in\{\lambda_1,\lambda_2\}$. The standard covariance
projection identity states that the variance after projection onto a unit
direction $\mathbf{u}$ is
$\mathbf{u}^\top\boldsymbol{\Sigma}_{2D,i}\mathbf{u}$.
Since $\mathbf{u}$ is also an eigenvector, this directional variance equals
$\lambda$. Applying the identity to \cref{eq:supp_cov_decomp} gives
\begin{align}
  \lambda
  &=
  \mathbf{u}^\top
  \boldsymbol{\Sigma}_{2D,i}
  \mathbf{u} \notag\\
  &=
  \mathbf{u}^\top
  \left(
    \sum_{d=1}^{3}
    s_d^2\mathbf{q}_d\mathbf{q}_d^\top
  \right)
  \mathbf{u} \notag\\
  &=
  \sum_{d=1}^{3}
  \underbrace{s_d^2
  \bigl(\mathbf{u}^\top\mathbf{q}_d\bigr)^2}_{w_d}.
  \label{eq:supp_directional_variance}
\end{align}
The last equality uses
$\mathbf{q}_d^\top\mathbf{u}=\mathbf{u}^\top\mathbf{q}_d$, since both are
the same scalar. 

Thus, $w_d$ measures the squared projected extent of the $d$-th scaled
3D axis along the observed overflow direction. Rather than selecting the
longest 3D axis for compression, ABR identifies the dominant axis by its
projected directional contribution:
\begin{equation}
  d^*=\arg\max_d w_d.
\end{equation}
Notably, the sign of $\mathbf{u}$ cancels in $w_d$, so opposite endpoints yield
the same axis attribution, although their boundary distances are evaluated
separately.

\subsection{Directional Compression and Multi-View Reconciliation}
\label{sec:supp_abr_compression}

Starting from $\boldsymbol{m}_i$, ABR approximates the first-exit distance
along the signed direction $\mathbf{u}$. Its ideal continuous definition is
\begin{equation}
  \label{eq:supp_directional_distance}
  \ell_{\mathbf{u}}
  =
  \inf\left\{
    t\geq0
    \,\middle|\,
    \boldsymbol{m}_i+t\mathbf{u}\notin\Omega
    \;\lor\;
    \mathcal{M}\!\left(
      \operatorname{round}(\boldsymbol{m}_i+t\mathbf{u})
    \right)=0
  \right\}.
\end{equation}
The implementation uniformly samples 129 points between the center and the
overflowing endpoint, rounds them to pixel locations, and places the estimated
boundary halfway between the last foreground sample and the first sample
outside the foreground or image. If no exit is observed before the endpoint,
the endpoint distance is used.

For each overflowing endpoint observation, ABR derives a candidate factor
for its attributed dominant axis. Recall from
\cref{eq:supp_directional_variance} that
$w_{d^*}=s_{d^*}^2
(\mathbf{u}^\top\mathbf{q}_{d^*})^2$.
During the local update, the original overflow direction $\mathbf{u}$ and
the first-order projection $\mathbf{q}_{d^*}$ of the unit 3D axis are held
fixed, only $s_{d^*}$ is changed to
$s'_{d^*}=f_{d^*}s_{d^*}$. Hence,
\begin{align}
  w'_{d^*}
  &=
  (s'_{d^*})^2
  (\mathbf{u}^\top\mathbf{q}_{d^*})^2 \notag\\
  &=
  (f_{d^*}s_{d^*})^2
  (\mathbf{u}^\top\mathbf{q}_{d^*})^2 \notag\\
  &=
  f_{d^*}^2w_{d^*}, \notag\\
  V'_{\mathbf{u}}
  &=
  \sum_{d\neq d^*}w_d+w'_{d^*} \notag\\
  &=
  \lambda-w_{d^*}+f_{d^*}^2w_{d^*}.
  \label{eq:supp_directional_update}
\end{align}

ABR represents the otherwise unbounded Gaussian support using the cutoff
$\sigma_c$. Along $\mathbf{u}$, the post-compression standard deviation is
$\sqrt{V'_{\mathbf{u}}}$, so the corresponding cutoff radius is
$\sigma_c\sqrt{V'_{\mathbf{u}}}$. To place this radius at the foreground
boundary, ABR matches it to the available directional distance
$\ell_{\mathbf{u}}$. Substituting \cref{eq:supp_directional_update} and
rearranging gives the complete derivation:
\begin{align}
  \sigma_c\sqrt{V'_{\mathbf{u}}}=\ell_{\mathbf{u}}
  &\Longleftrightarrow
  V'_{\mathbf{u}}
  =
  \left(\frac{\ell_{\mathbf{u}}}{\sigma_c}\right)^2, \notag\\
  \lambda-w_{d^*}+f_{d^*}^2w_{d^*}
  &=
  \left(\frac{\ell_{\mathbf{u}}}{\sigma_c}\right)^2, \notag\\
  f_{d^*}^2w_{d^*}
  &=
  \left(\frac{\ell_{\mathbf{u}}}{\sigma_c}\right)^2
  -\lambda+w_{d^*}, \notag\\
  f_{d^*}^2
  &=
  \frac{
    (\ell_{\mathbf{u}}/\sigma_c)^2
    -\lambda+w_{d^*}
  }{w_{d^*}}, \notag\\
  f_{d^*}
  &=
  \sqrt{
    \frac{
      (\ell_{\mathbf{u}}/\sigma_c)^2
      -\lambda+w_{d^*}
    }{w_{d^*}}
  }.
  \label{eq:supp_compression_factor}
\end{align}
This is the compression rule in Eq.~(13) of the main paper, with the
intermediate variance balance made explicit. Here, $\lambda$, $w_{d^*}$, and
$(\ell_{\mathbf{u}}/\sigma_c)^2$ are all in pixel-squared units, so
$f_{d^*}$ is dimensionless and the positive root is used.

Each overflowing endpoint $\boldsymbol{e}$ in view $j$ yields a single-view
candidate factor $f_{i,d}^{(j,\boldsymbol{e})}$ for its attributed dominant
axis $d$. Because different observations of the same Gaussian may attribute
the overflow to different axes, ABR reconciles these candidates separately
for every ``(Gaussian, axis)'' pair. Specifically, it takes the smallest
factor supported by the corresponding endpoints and views, and clips the
result to $[f_{\min},1]$:
\begin{equation}
  \label{eq:supp_multiview_factor}
  f_{i,d}^{\mathrm{final}}
  =
  \min\!\left(
    1,\,
    \max\!\left(
      f_{\min},
      \min_{(j,\boldsymbol{e})\in\mathcal{A}_{i,d}}
      f_{i,d}^{(j,\boldsymbol{e})}
    \right)
  \right),
\end{equation}
where $\mathcal{A}_{i,d}$ contains the overflow observations attributed to
axis $d$ of $g_i$, and $f_{\min}$ prevents excessive compression. The minimum
implements the most restrictive supported correction, while the interval
prevents both scale enlargement and excessive shrinkage.

For the single-view solution, a factor above one is replaced by one, and a
positive factor below $f_{\min}$ is replaced by $f_{\min}$. If
$(\ell_{\mathbf{u}}/\sigma_c)^2\leq\lambda-w_{d^*}$, the fixed contribution
of the other two axes already reaches or exceeds the desired variance, so no
admissible factor in $[f_{\min},1]$ can attain the boundary exactly and ABR then
uses $f_{\min}$.

Notably, a reconciled factor is applied only when the overflow has sufficient
cross-view support. ABR measures this support by
\begin{equation}
  \gamma_i
  =
  \frac{n_i^{\mathrm{ovf}}}{n_i^{\mathrm{vis}}},
\end{equation}
where $n_i^{\mathrm{vis}}$ counts eligible center-in-foreground observations
of $g_i$, and $n_i^{\mathrm{ovf}}$ counts those with at least one overflowing
endpoint. The counts use aligned training views with retained masks from the
latest completed segmentation round; missing masks and supplemental spherical
views are excluded. ABR applies the reconciled update only when
$\gamma_i>\rho$:
\begin{equation}
  \log s_{i,d}
  \leftarrow
  \log s_{i,d}+\log f_{i,d}^{\mathrm{final}}.
\end{equation}
This cross-view consensus gate improves robustness to isolated mask errors.
A poorly segmented view may produce both a false overflow detection and an
unreliable candidate factor, but it cannot by itself trigger a 3D scale
update unless the overflow is corroborated by a sufficient fraction of
valid views.
\section{Extended Experimental Evaluation}
\label{sec:supp_experiments}

\subsection{Per-Stage Runtime Analysis}
\label{sec:supp_timing}

\Cref{tab:timing} reports the timing breakdown of each stage in the VCAR
inference pipeline. All timings are measured on a single NVIDIA A100 GPU
and averaged over all objects in each dataset.

The major computational costs are concentrated in rendering, SAM
segmentation, and voting on the larger LERF scenes. Notably, coarse-stage
rendering on LERF is significantly more expensive than fine-stage rendering.
This is primarily a first-pass overhead: the coarse stage performs initial
loading of camera poses and camera metadata before rendering begins, whereas
later stages reuse cached camera information. This explains why coarse
rendering takes approximately 25 seconds on LERF while fine rendering takes
only about 5 seconds.

For NVOS, all non-SAM modules remain lightweight, with rendering, voting,
view completeness assessment (VCA), and final rendering each taking around
2--3 seconds. For LERF, SAM segmentation remains the dominant component,
taking about 25 seconds in the coarse stage and about 30 seconds in the fine
stage. ABR takes approximately 3 seconds on NVOS and 10 seconds on LERF,
while final rendering after ABR adds about 2 seconds and 5 seconds,
respectively.

In summary, VCAR's runtime is dominated by 2D mask generation (SAM), while
VCA and ABR introduce moderate overhead. With conditional SSS triggering and
cache reuse across stages, the full training-free pipeline remains practical
at $\sim$30\,s per object on NVOS and $\sim$120\,s on LERF.

\begin{table}[t]
  \caption{Per-stage timing breakdown (seconds) of VCAR on NVOS and LERF.
    ``VCA'' denotes view completeness assessment.
    Objects that do not require SSS skip the fine-stage rendering and SAM
    segmentation, resulting in shorter total time.
    ``Final rendering'' denotes the rendering stage after ABR refinement.}
  \label{tab:timing}
  \centering
  \small
  \begin{tabular}{lcc}
    \toprule
    Stage                             & NVOS (s)          & LERF (s)           \\
    \midrule
    Coarse rendering                  & $\sim$2           & $\sim$25           \\
    Coarse SAM segmentation           & $\sim$5           & $\sim$25           \\
    Coarse voting                     & $\sim$2           & $\sim$10           \\
    VCA                               & $\sim$2           & $\sim$5            \\
    \midrule
    Fine rendering (if SSS triggered) & $\sim$2           & $\sim$5            \\
    Fine SAM segmentation             & $\sim$10          & $\sim$30           \\
    Fine voting                       & $\sim$2           & $\sim$5            \\
    \midrule
    ABR                               & $\sim$3           & $\sim$10           \\
    Final rendering                   & $\sim$2           & $\sim$5            \\
    \midrule
    \textbf{Total}                    & $\sim$\textbf{30} & $\sim$\textbf{120} \\
    \bottomrule
  \end{tabular}
\end{table}

\subsection{Backbone Portability with SAM 2}
\label{sec:supp_backbone}

The main paper reports VCAR results with SAM\,3 \cite{SAM3} as the default 2D backbone.
To validate portability, we run VCAR with SAM\,2 \cite{ravi2024sam} on LERF using the same pipeline and hyperparameter settings.
This section jointly summarizes 2D mask stability
and end-to-end 3D segmentation performance under SAM\,2.

\Cref{tab:sam2_2d} reports frame-wise PSNR/SSIM between SAM\,2 and
SAM\,3 masks from the same rendered frames. Their overall agreement reaches
97.0\% SSIM and 28.5\,dB PSNR, with the lowest agreement occurring in the
more cluttered \textit{kitchen} scene.

\Cref{tab:sam2_3d} reports end-to-end 3D segmentation results on LERF with SAM\,2 or SAM\,3, under identical pipeline settings, together with their performance gaps.
VCAR with SAM\,2 remains competitive
with 72.7\% average mIoU on LERF.
The largest degradation appears in \textit{kitchen} (\,$\Delta$mIoU $=-6.4$), where many small objects are visible only in limited views, making both temporal mask continuity and downstream 3D decisions more sensitive to 2D errors.
Even in this most challenging case,
VCAR+SAM\,2 reaches 68.3\% mIoU,
which remains above the previous best baseline LangSplatV2 (59.1\%) reported in the main paper.

Overall, replacing SAM\,3 with SAM\,2 changes average mIoU by 2.4 points
and mAcc by 0.4 points. The relatively small average gap indicates that the
VCAR pipeline transfers beyond its default 2D backbone, while the larger
\textit{kitchen} difference also shows that mask quality remains relevant
in difficult scenes. We use SAM\,3 in the main paper because it achieves
the stronger average result and directly supports text prompts.

\begin{table}[t]
  \caption{2D mask similarity between masks generated by SAM 2 and SAM 3.}
  \label{tab:sam2_2d}
  \centering
  {\footnotesize
    \setlength{\tabcolsep}{3.5pt}
    \begin{tabular}{lccccc}
      \toprule
      Metric    & Figurines & Ramen & Teatime & Kitchen & Overall \\
      \midrule
      PSNR (dB) & 33.8      & 30.7  & 29.6    & 20.0    & 28.5    \\
      SSIM (\%) & 99.6      & 99.3  & 96.8    & 92.3    & 97.0    \\
      \bottomrule
    \end{tabular}
  }
\end{table}

\begin{table}[t]
  \caption{End-to-end 3D segmentation results on LERF with SAM\,2 or SAM\,3.
    $\Delta$mAcc and $\Delta$mIoU denote the differences (``SAM\,2 $-$ SAM\,3'') in mAcc and mIoU, respectively.}
  \label{tab:sam2_3d}
  \centering
  {\footnotesize
    \setlength{\tabcolsep}{3.5pt}
    \begin{tabular}{lccccc}
      \toprule
      Metric           & Figurines & Ramen & Teatime & Kitchen & Overall \\
      \midrule
      SAM\,2 mIoU (\%) & 71.7      & 68.3  & 82.3    & 68.3    & 72.7    \\
      SAM\,3 mIoU (\%) & 73.8      & 68.4  & 83.4    & 74.7    & 75.1    \\
      $\Delta$mIoU     & -2.1      & -0.1  & -1.1    & -6.4    & -2.4    \\
      \midrule
      SAM\,2 mAcc (\%) & 99.5      & 98.9  & 99.4    & 96.7    & 98.6    \\
      SAM\,3 mAcc (\%) & 99.6      & 98.9  & 99.4    & 98.1    & 99.0    \\
      $\Delta$mAcc     & -0.1      & 0.0   & 0.0     & -1.4    & -0.4    \\
      \bottomrule
    \end{tabular}
  }
\end{table}

\setcounter{figure}{0}
\begin{figure}[!t]
  \centering
  \includegraphics[width=0.8\columnwidth]{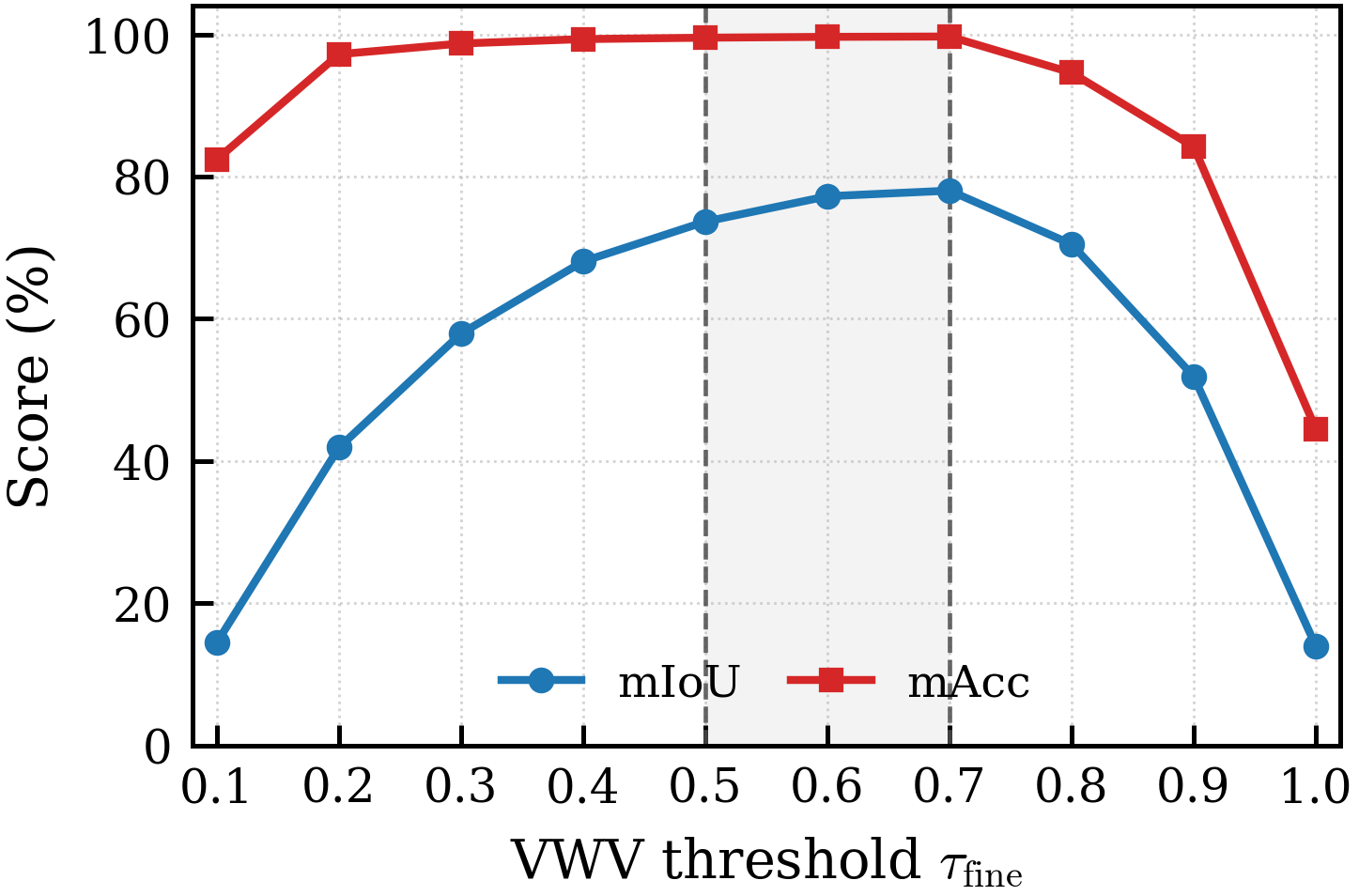}
  \caption{Sensitivity to the fine-stage VWV threshold
    $\tau_{\mathrm{fine}}$ on LERF. Dashed lines mark the selected interval
    $[0.5,0.7]$.}
  \Description{A line chart of mIoU and mAcc as the fine-stage voting
    threshold varies from 0.1 to 1.0. Both metrics are strongest in the
    middle range and decrease at extreme thresholds.}
  \label{fig:supp_threshold}
\end{figure}

\subsection{Fine-Stage Voting Threshold Sensitivity}
\label{sec:supp_threshold}

\Cref{fig:supp_threshold} sweeps $\tau_{\mathrm{fine}}$ while holding the
remaining pipeline fixed. A low threshold accepts Gaussians supported by
only a small fraction of visible views and therefore admits more background
evidence. Conversely, a high threshold requires near-unanimous support and
can reject target Gaussians that are visible or correctly segmented in only
part of the view set. The resulting mIoU is highest in the intermediate
range, while mAcc also remains near its maximum there. We consequently use
$\tau_{\mathrm{fine}}\in[0.5,0.7]$ for LERF, as reported in the main paper,
to balance target completeness and background suppression. This sweep
characterizes the threshold choice for the evaluated LERF cases and should
not be interpreted as a universal setting for arbitrary scenes.


\begin{table}[!t]
  \caption{Boundary-aligned evaluation (\%) on the eight LERF objects that
    trigger SSS. ``Coarse Only'' is the train-view SAM\,3 + VWV baseline.
    The +SSS and +ABR rows independently add the indicated module to that
    baseline; Full includes both modules.}
  \label{tab:supp_boundary_metrics}
  \centering
  {\footnotesize
    \setlength{\tabcolsep}{5pt}
    \begin{tabular}{lccccc}
      \toprule
      Configuration & mIoU & mAcc & B-F1 & B-IoU & T-IoU \\
      \midrule
      Coarse Only & 53.4 & 90.2 & 32.4 & 18.8 & 54.7 \\
      +SSS        & 75.0 & 96.9 & 56.1 & 34.1 & 59.0 \\
      +ABR        & 77.6 & 97.3 & 65.0 & 43.1 & 60.0 \\
      Full        & \textbf{82.2} & \textbf{97.8} & \textbf{69.5}
                  & \textbf{45.6} & \textbf{62.5} \\
      \bottomrule
    \end{tabular}
  }
\end{table}

\subsection{Boundary-Aligned Evaluation on Selected LERF Objects}
\label{sec:supp_boundary_metrics}

Because SSS and ABR mainly address incomplete or overflowing boundaries,
\cref{tab:supp_boundary_metrics} complements the region-level ablation with
B-F1, B-IoU, and T-IoU. We evaluate the eight LERF objects that satisfy
$\Delta_{\max}>\Delta_{\mathrm{th}}$ and therefore trigger SSS:
\textit{Figurines--Apple}, \textit{Figurines--Camera},
\textit{Ramen--Bowl}, \textit{Ramen--Sake Cup},
\textit{Teatime--Sheep}, \textit{Teatime--Bear},
\textit{Kitchen--Fridge}, and \textit{Kitchen--Toaster}.
The reported averages therefore characterize this fixed, method-defined
subset rather than the complete LERF benchmark. All configurations follow
the same evaluation and aggregation protocol.

Both +SSS and +ABR improve all three boundary metrics over Coarse Only, and
Full achieves the best result on every reported metric. This pattern is
consistent with their distinct roles: SSS adds evidence from under-observed
directions, whereas ABR reduces projected overflow.

\setcounter{figure}{1}
\begin{figure*}[!t]
  \centering
  \includegraphics[width=0.97\textwidth]{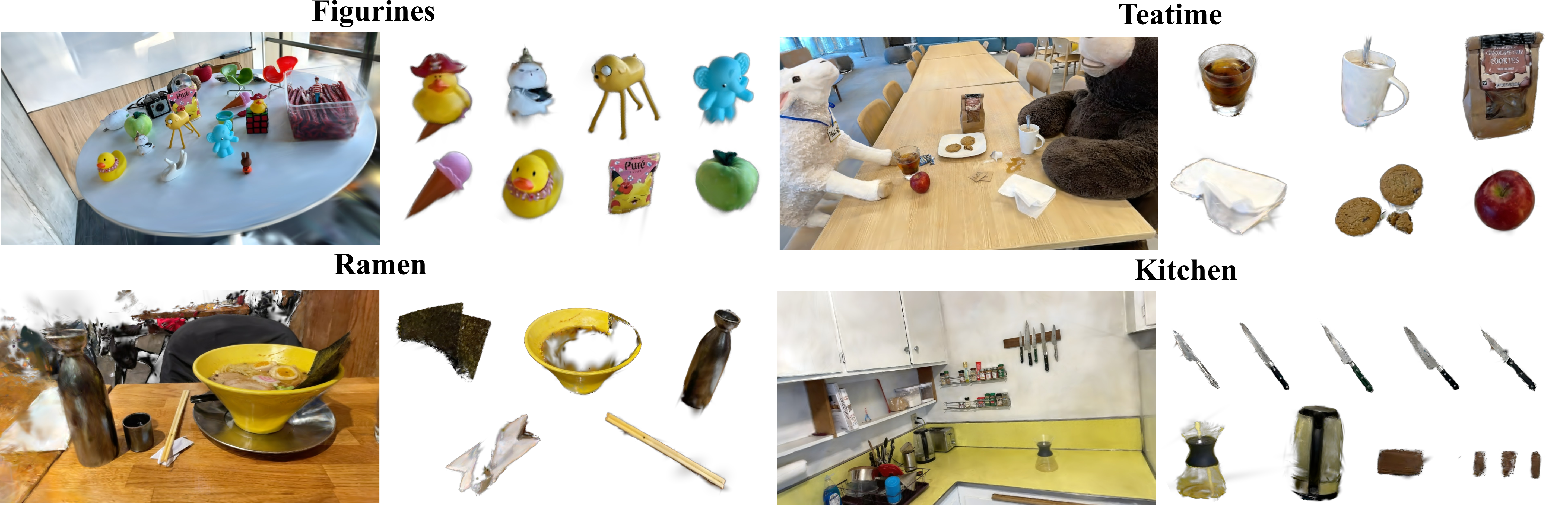}
  \caption{Additional qualitative comparisons on LERF. Across the shown
    examples, VCAR yields cleaner boundaries and fewer floating artifacts
    under cluttered layouts and inter-object occlusion.}
  \Description{Additional qualitative comparisons on the LERF benchmark.}
  \label{fig:supp_vis_lerf}
\end{figure*}

\begin{figure*}[!t]
  \centering
  \includegraphics[width=0.97\textwidth]{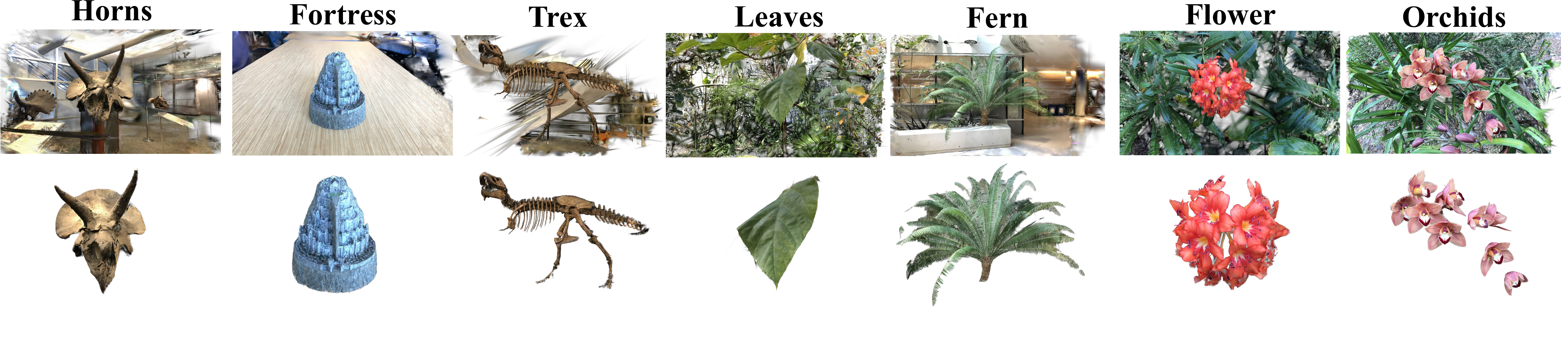}
  \caption{Additional qualitative comparisons on NVOS. Across the shown
    examples, VCAR improves boundary completeness in under-observed regions
    while reducing background leakage in forward-facing captures.}
  \Description{Additional qualitative comparisons on the NVOS benchmark.}
  \label{fig:supp_vis_nvos}
\end{figure*}

\subsection{Additional Qualitative Comparisons}
\label{sec:supp_visualization}

We provide additional qualitative comparisons beyond those included in the
main paper. In the LERF examples in \cref{fig:supp_vis_lerf}, VCAR produces
cleaner object boundaries and fewer floating fragments under clutter and
inter-object occlusion. In the shown NVOS examples in
\cref{fig:supp_vis_nvos}, the supplementary viewpoints improve the recovery
of under-observed regions while ABR suppresses part of the boundary leakage.
These results complement the quantitative evaluation across objects of
different sizes and capture layouts.

\section{Additional Clarifications}
\label{sec:supp_clarifications}

\subsection{Robustness to Imperfect 2D Masks}
\label{sec:supp_mask_robustness}

\begin{figure}[H]
  \centering
  \begingroup
  \setlength{\tabcolsep}{0.7pt}
  \begin{tabular}{@{}cccc@{}}
    \includegraphics[width=1.0\columnwidth]{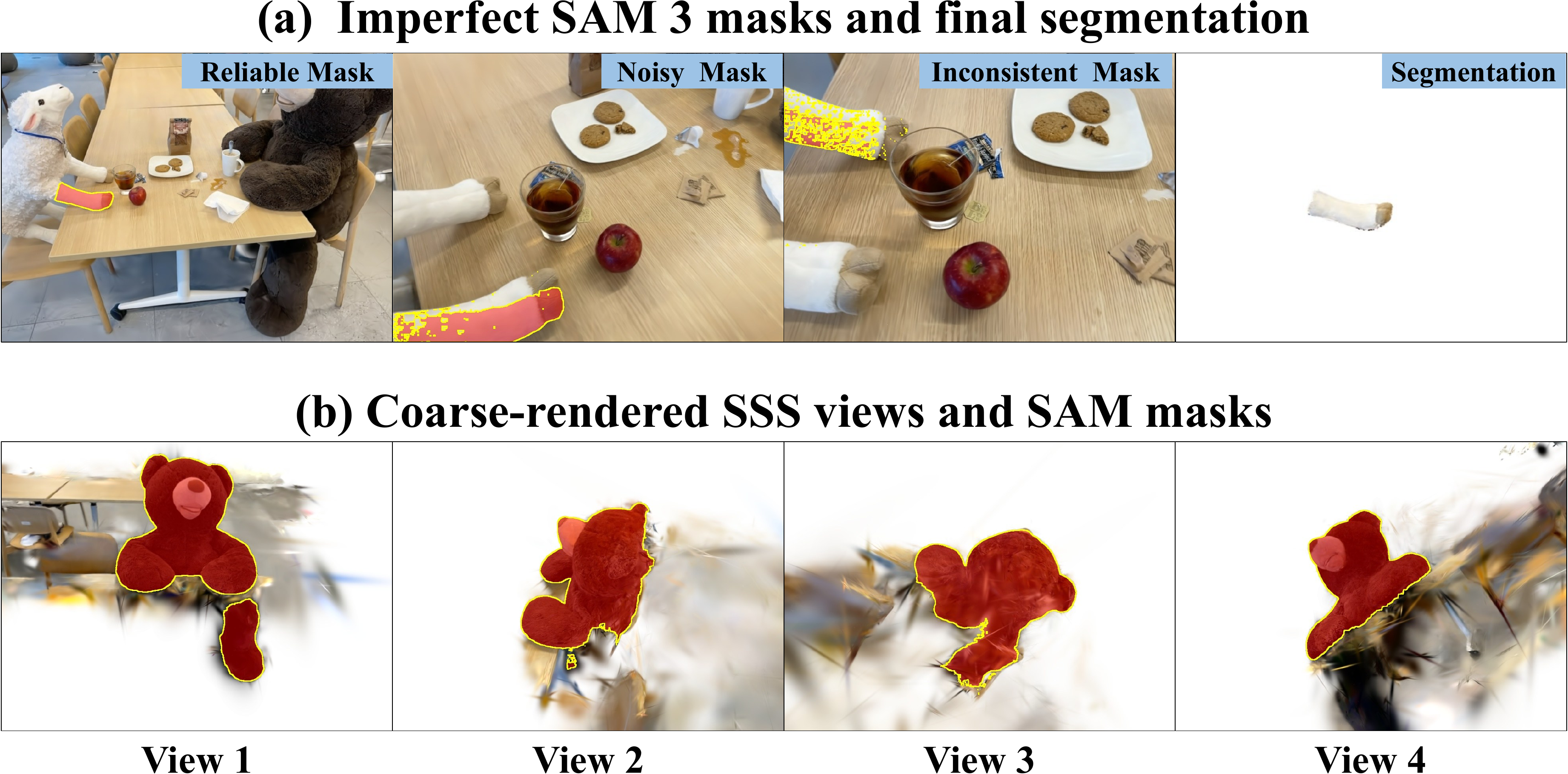}
  \end{tabular}
  \endgroup
  \caption{Robustness to imperfect masks and examples of SSS-generated views.
  (a) Three representative SAM\,3 masks of varying quality and the resulting
  3D segmentation. In this example, isolated noisy or inconsistent masks are
  suppressed by multi-view aggregation, yielding a stable result.
  (b) Four supplementary views rendered from
  $\mathcal{G}^{\mathrm{coarse}}$ and used to generate additional 2D masks.
  These intermediate views may contain coarse-rendering artifacts.}
  \Description{The top row shows reliable, noisy, and inconsistent SAM\,3
    masks followed by the resulting 3D segmentation. The bottom row shows
    four coarse-rendered supplementary views and their masks.}
  \label{fig:supp_robustness_sss}
\end{figure}

VCAR uses pretrained 2D masks as evidence and therefore cannot be
independent of their quality. Within a SAM\,3 propagation, some mask errors
are sporadic: they affect only a small number of views while most other
views provide consistent foreground evidence. VWV normalizes votes over
views in which a Gaussian is visible, so an isolated incorrect mask
contributes only a limited fraction of the final score and can be diluted
by the remaining views. \Cref{fig:supp_robustness_sss}(a) illustrates this
case, where noisy or inconsistent masks coexist with a stable final
segmentation.

This robustness is conditional rather than absolute. If the same semantic
error persists across many views, the incorrect evidence becomes the
multi-view consensus and is inherited by VWV. SSS can add viewpoints but
does not correct a repeatedly wrong mask, and ABR only modifies the scale
of already selected Gaussians rather than their semantic label. Persistent
2D errors therefore remain an upstream limitation of the complete pipeline.

\subsection{Role and Quality of SSS-Generated Views}
\label{sec:supp_sss_quality}

SSS-generated views are intermediate observations rather than
photorealistic novel-view synthesis outputs. They are rendered using
$\mathcal{G}^{\mathrm{coarse}}$, so missing background content and coarse
object geometry can produce unnatural appearance or rendering artifacts,
as shown in \cref{fig:supp_robustness_sss}(b). Moreover, a sampled camera
usually has no corresponding ground-truth image, making image-level PSNR
poorly aligned with the purpose of SSS. The relevant criterion is whether a
sampled view reduces background occlusion, exposes a useful object-facing
direction, and provides an additional mask that improves the downstream 3D
segmentation. The ablation gains of +SSS and Full in
\cref{tab:supp_boundary_metrics} quantify this downstream utility.

\begin{figure*}[!t]
  \centering
  \includegraphics[width=0.82\textwidth]{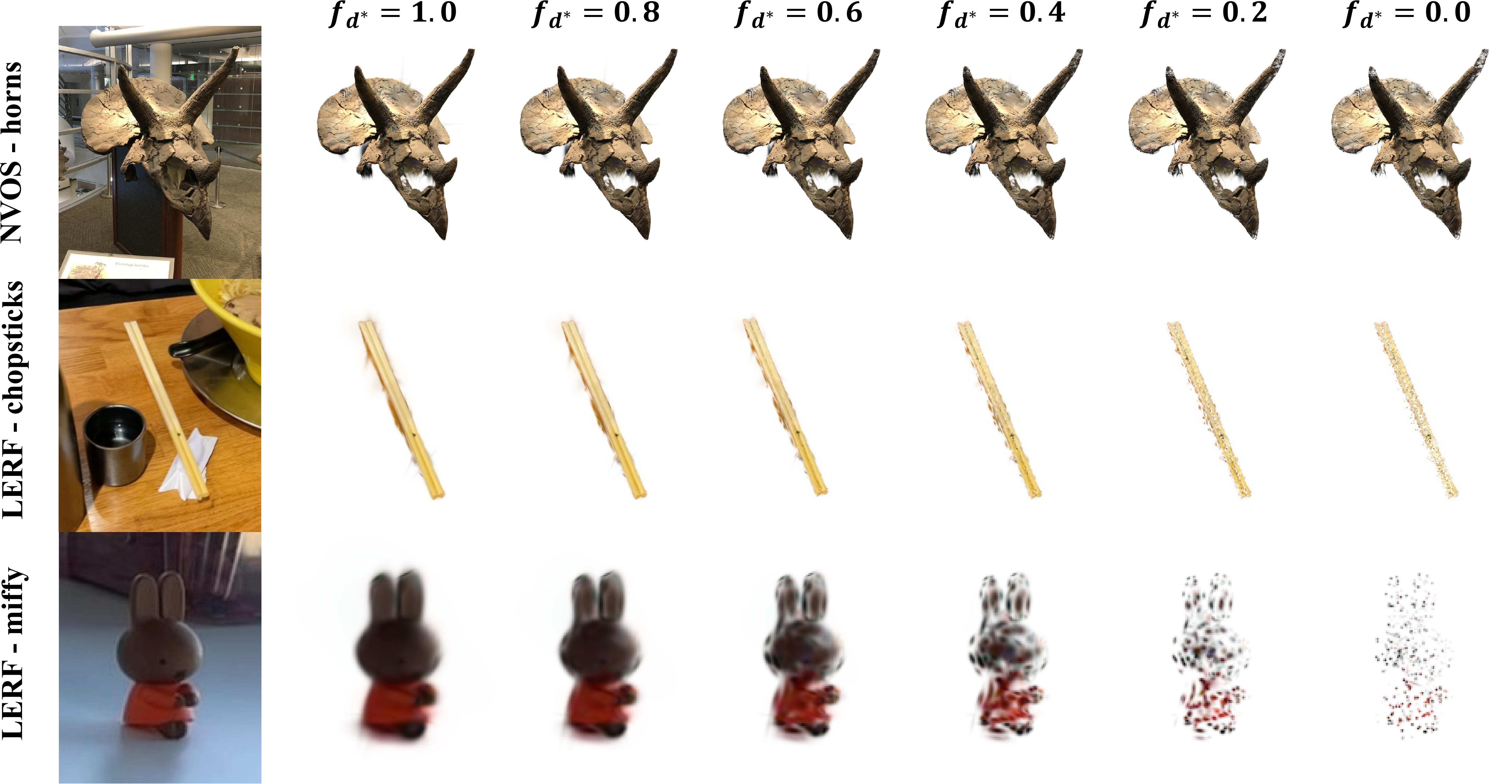}
  \caption{Diagnostic sweep of boundary-axis compression. The leftmost
    column shows the reference images. The remaining columns apply a fixed
    factor $f_{d^*}\in\{1.0,0.8,0.6,0.4,0.2,0.0\}$ to every Gaussian-axis
    pair attributed to an overflow in at least one view. Here,
    $f_{d^*}=1.0$ leaves the selected axis unchanged, whereas
    $f_{d^*}=0.0$ completely compresses it.}
  \Description{A three-row comparison of a large horned skull, a pair of
    chopsticks, and a small rabbit figure under progressively stronger
    boundary-axis compression from 1.0 to 0.0. Strong compression causes
    contour erosion, speckled artifacts, and severe fragmentation,
    respectively.}
  \label{fig:supp_abr_sweep}
\end{figure*}

\subsection{ABR versus Visual Quality}
\label{sec:supp_abr_sweep}

To isolate the effect of compression strength, we apply a common factor
$f_{d^*}$ to every Gaussian-axis pair attributed to an overflow in at least
one view, instead of using ABR's adaptive factors.
\Cref{fig:supp_abr_sweep} varies $f_{d^*}$ from no compression
($f_{d^*}=1.0$) to complete compression ($f_{d^*}=0.0$), revealing
scale- and geometry-dependent effects. For large objects such as
\textit{Horns}, ABR compressions accumulates over the dense, relatively
small boundary Gaussians and progressively erodes the outer contour. For
elongated or irregular structures such as \textit{Chopsticks}, comparatively
large Gaussians support the thin shape, so compressing them disrupts spatial
continuity and produces speckled gaps. For small objects such as
\textit{Miffy}, only a few Gaussians represent the object, and a single
Gaussian may cover both interior and boundary regions, therefore, its compression 
probably can remove object content and cause severe fragmentation.

Overall, a geometrically tighter boundary does not necessarily produce a
visually better rendering. An anisotropic Gaussian may have a long projected
tail that crosses the foreground boundary yet remains visually
inconspicuous after alpha compositing. Compressing its attributed axis can therefore
suppress visually meaningful Gaussian contributions to the rendered object, 
exposing the trade-off between reducing geometric overflow with ABR and preserving visual fidelity.

\subsection{Coarse-Stage Coverage and Occlusion-Induced Failures}
\label{sec:supp_coverage_failure}

The coarse set $\mathcal{G}^{\mathrm{coarse}}$ is not theoretically
guaranteed to contain every target Gaussian. SSS renders supplementary views
from that coarse set, and ABR only adjusts the scale of selected Gaussians.
Consequently, if a target part is absent after coarse voting, neither module
can recover it from scratch. The direct effect is reduced recall for thin,
weakly visible, or heavily occluded parts.

The missing hind leg of \textit{Teatime--Sheep} is one such case. Tables and
chairs occlude the part in most original views, causing the corresponding
SAM\,3 masks to omit it repeatedly. When incomplete masks dominate the
visible-view evidence, VWV excludes the associated Gaussians from the
coarse result. The later SSS and ABR stages then have no foreground support
from which to restore the leg. This is an upstream coverage failure caused by
persistent omission rather than a failure of the later refinement stages.

\end{document}